\documentclass[nohyperref]{article}

\usepackage{microtype}
\usepackage{graphicx}
\usepackage{subfigure}
\usepackage{booktabs} % for professional tables

\usepackage{hyperref}

\usepackage[accepted]{icml2022}

\usepackage{amsmath}
\usepackage{amssymb}
\usepackage{mathtools}
\usepackage{amsthm}
\usepackage{multirow}
\usepackage{footmisc}
\usepackage[capitalize,noabbrev]{cleveref}

\theoremstyle{plain}

\theoremstyle{definition}

\theoremstyle{remark}

\usepackage[textsize=tiny]{todonotes}

\icmltitlerunning{Towards Theoretical Analysis of Transformation Complexity of ReLU DNNs}

\begin{document}

\twocolumn[
\icmltitle{Towards Theoretical Analysis of Transformation Complexity of ReLU DNNs}

\icmlsetsymbol{equal}{*}

\begin{icmlauthorlist}
\icmlauthor{Jie Ren}{equal,sjtu}
\icmlauthor{Mingjie Li}{equal,sjtu}
\icmlauthor{Meng Zhou}{cmu}
\icmlauthor{Shih-Han Chan}{ucsd}
\icmlauthor{Quanshi Zhang}{zqs}
\end{icmlauthorlist}

\icmlaffiliation{sjtu}{Shanghai Jiao Tong University.}
\icmlaffiliation{cmu}{Carnegie Mellon University.}
\icmlaffiliation{ucsd}{University of California San Diego.}
\icmlaffiliation{zqs}{Quanshi Zhang is the corresponding author. He is with the Department of Computer Science and Engineering,
the John Hopcroft Center, and the MoE Key Lab of Artificial Intelligence, AI Institute, at the Shanghai Jiao Tong University, China}

\icmlcorrespondingauthor{Quanshi Zhang}{zqs1022@sjtu.edu.cn}

\vskip 0.3in
]



\printAffiliationsAndNotice{\icmlEqualContribution} % otherwise use the standard text.

\begin{abstract}
This paper aims to theoretically analyze the complexity of feature transformations encoded in piecewise linear DNNs with ReLU layers. We propose metrics to measure three types of complexities of transformations based on the information theory. We further discover and prove the strong correlation between the complexity and the disentanglement of transformations. Based on the proposed metrics, we analyze two typical phenomena of the change of the transformation complexity during the training process, and explore the ceiling of a DNN's complexity. The proposed metrics can also be used as a loss to learn a DNN with the minimum complexity, which also controls the over-fitting level of the DNN and influences adversarial robustness, adversarial transferability, and knowledge consistency.
Comprehensive comparative studies have provided new perspectives to understand the DNN. 
The code is released at \href{https://github.com/sjtu-XAI-lab/transformation-complexity}{\textit{https://github.com/sjtu-XAI-lab/transformation-complexity}}.
\end{abstract}

\section{Introduction}
\label{sec:introduction}
Understanding the black-box of deep neural networks (DNNs) has attracted increasing attention in recent years. Previous studies usually interpret DNNs by either explaining DNNs visually/semantically~\cite{lundberg2017unified,ribeiro2016should}, or analyzing the feature representation capacity of a DNN~\cite{higgins2017beta,achille2018emergence,achille2018information,fort2019stiffness}.
To this end, in this paper, we aim to explain the representation capacity of a DNN by analyzing its complexity of feature representations.

\begin{figure}[t]
    \centering
    \includegraphics[width=\linewidth]{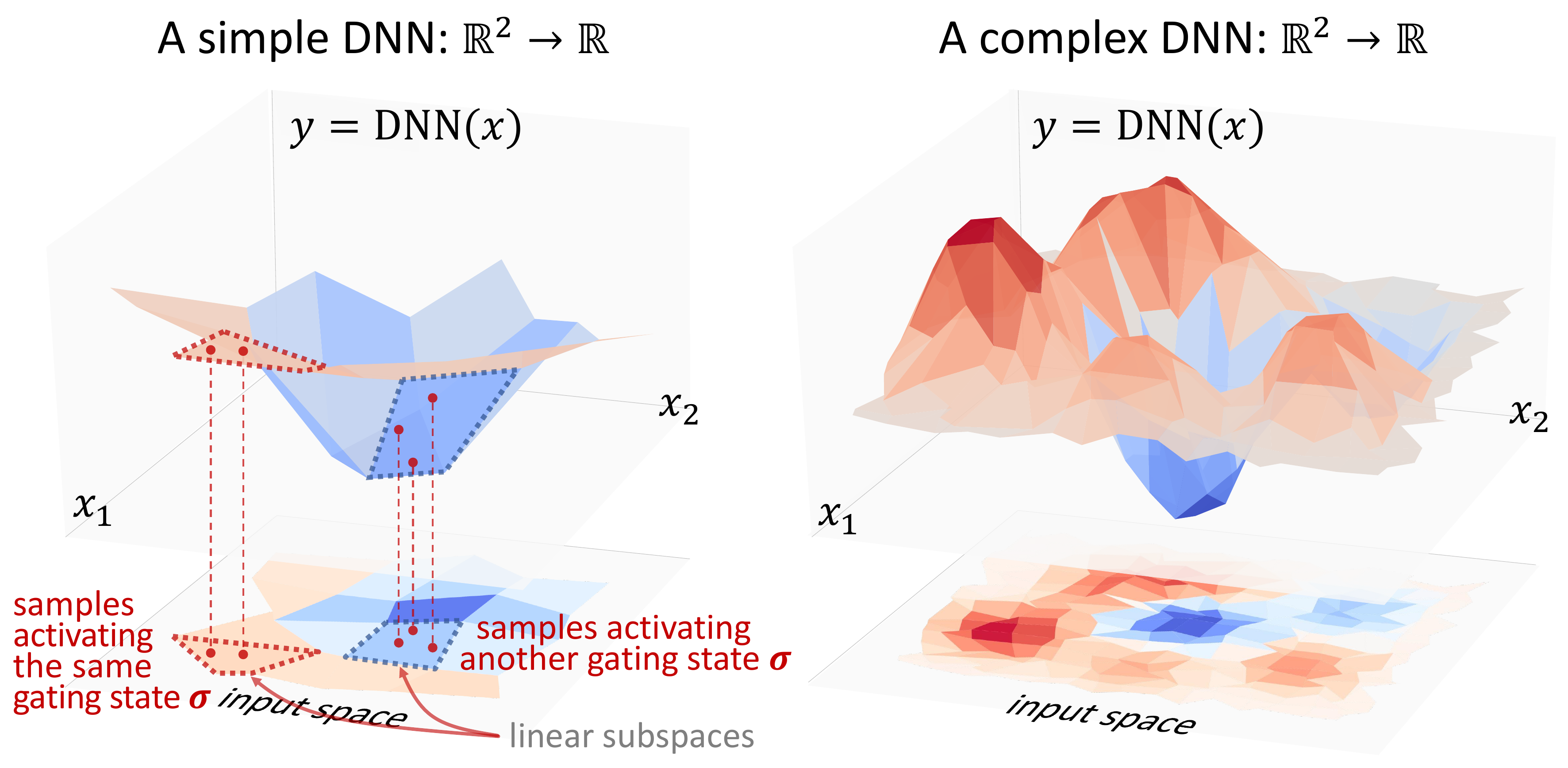}
    \vspace{-15pt}
    \caption{The change of gating states in an ReLU DNN divides the input space into lots of linear subspaces. In this paper, we aim to quantify the complexity of linear subspaces encoded in ReLU DNNs.}
    \label{fig:intro}
\end{figure}

In fact, there have been many studies~\cite{arora2016understanding, zhang2016architectural, raghu2017expressive, manurangsi2018computational} on the representation complexity of a DNN. However, unlike traditional studies on the theoretic maximum complexity of a DNN, we investigate this problem from a new perspective, \emph{i.e.}, the complexity of piecewise linear representations of a ReLU network. In fact, \textbf{most DNNs with ReLU activation functions are piecewise linear models. As Figure~\ref{fig:intro} shows, for such piecewise linear models, the number of piecewise linear subspaces in the model is the distinctive perspective to analyze the representation complexity of the model.} A complex DNN usually uses countless linear subspaces for regression/classification. Such a complexity reveals new insights into typical problems, such as the feature disentanglement, over-fitting, adversarial robustness, adversarial transferability, and knowledge consistency.

Therefore, in this paper, we define three new metrics in information theory to quantify the diversity of gating states in gating layers (\emph{e.g.}, ReLU, max-Pooling, and Dropout layers). Such gating states directly determine the complexity of linear subspaces that are encoded by a piecewise linear ReLU network.
For example, as Figure~\ref{fig:intro} shows, the change of gating states in a DNN divides the entire input space into lots of linear subspaces. Compared with intermediate-layers neural activations, gating states can better reflect the diversity of transformations in a piecewise linear DNN.

Thus, the first metric is defined as the entropy of the gating states of nonlinear operations.
Specifically, let us consider the task {$(X,Y)$}, where {$X$} denotes all input samples and {$Y$} denotes corresponding labels. Given a DNN, the complexity of transformations represents the diversity of transformations that map each input {$x$} to the corresponding {$y\in Y$} over all inputs {$x \in X$}, \emph{i.e.,} the number of linear subspaces in piecewise linear transformations.
From this perspective, the simplest model is a linear transformation {$y=f(w^{\textrm{T}}x+b)=w^{\textrm{T}}x+b$}, where the nonlinear operation {$f(\cdot)$} is by-passed as an all-passing gating layer.
In this case, the entropy of its gating states is zero.
In comparison, as Figure~\ref{fig:intro} shows, a DNN usually generates various gating states for different inputs, thereby dividing all training samples into different subspaces.
The complexity of such a piecewise linear representation is quantified as the entropy of gating states of all gating layers. 
Let the binary vector {$\sigma_{l}=[\sigma_{l,1},\sigma_{l,2},\dots,\sigma_{l,D}]\in\{0,1\}^{D}$} denote gating states of the gating layer.
Let $\Sigma$ represent the random variable of all gating states in all layers of the DNN.
In this way, the entropy {$H(\Sigma)$} among all inputs measures the complexity of the overall transformation complexity in the DNN.

Based on the entropy {$H(\Sigma)$}, we further propose {$I(X;\Sigma)$} and  {$I(X;\Sigma;Y)$} as two additional metrics to further disentangle specific components of the overall transformation complexity {$H(\Sigma)$} in a fine-grained manner. The mutual information {$I(X;\Sigma)$} measures the complexity of transformations that are caused by the input. 
The mutual information {$I(X;\Sigma;Y)$} represents the complexity of transformations that are caused by the input and are directly used for inference. For example, in the task of object classification, not all transformations in the DNN are category-specific. {$I(X;\Sigma;Y)$} reflects category-specific components of transformations.
Notice that {$I(X;\Sigma)\!=\! H(\Sigma)$} in most cases.
However, for other cases when the DNN uses additional transformations (randomness) beyond transformations caused by the input {$X$}, we have {$I(X;\Sigma)\!\ne\! H(\Sigma)$}.
For example, sampling operations in the VAE~\cite{kingma2013auto} and the dropout operation bring additional transformations.

\textbf{Analysis and explanations.} In this paper, we prove various properties of the complexity metrics.
Let us focus on the complexity of transforming the feature in an intermediate layer to the output of the DNN.
The transformation complexity decreases through the layerwise propagation.
In other words, deep features usually require simpler transformations to conduct inference than shallow features.
Then, we use the proposed complexity metrics to analyze the knowledge representation of DNNs.
We can summarize the change of the complexity during the training process into two types.
In traditional stacked DNNs without skip-connections, the complexity usually decreases first, and increases later. This indicates that DNNs may drop noisy features in early stages, and learn useful information later. 
Whereas, in residual networks, the transformation complexity does not decrease during the learning process.

In this study, we conduct the following theoretic explorations based on the complexity metrics.

(1) \textbf{Disentanglement: we prove the strong correlation between the complexity and the disentanglement of transformations.}
Let us consider DNNs with similar activation rates in a certain layer. If the complexity of the transformation encoded in a specific layer is larger, then gating states of different feature dimensions tend to be more independent with each other.

(2)  \textbf{Minimum complexity and the gap between the training loss and the testing loss: we use the complexity as a loss to learn a DNN with the minimum complexity.}
A DNN usually learns over-complex transformations \emph{w.r.t.} the task. Thus, we propose a complexity loss, which penalizes unnecessary transformations to learn a DNN with the minimum complexity.
Given the DNN learned using the complexity loss, we find that the gap between the training loss and the testing loss decreases, when we reduce the complexity of transformations.
Furthermore, we find that the complexity loss also influences adversarial robustness, adversarial transferability, and knowledge consisitency.

(3) \textbf{Maximum complexity: we explore the ceiling of a DNN's complexity.}
\textit{
First, the complexity of a DNN does not monotonously increase with the network depth.}
In contrast, the traditional stacked DNN with a deep architecture may encode simpler transformations than shallow DNNs in some cases.
Whereas, the transformation complexity of residual networks is saturated when we use more gating layers. \textit{Second, the complexity of transformations does not increase monotonously along with the increase of the complexity of tasks.} In contrast, if the task complexity exceeds a certain limit, the complexity of transformations encoded in the DNN will decrease along with the increase of the task complexity.

Theorectial contributions of this study can be summarized as follows. (1) We define three metrics to evaluate the complexity of transformations in piecewise linear DNNs, which have a great theoretical extensibility. (2) We prove the strong correlation between the complexity and the disentanglement of transformations. (3) We further use the transformation complexity as a loss to learn a minimum-complexity DNN, which also reduces the gap between the training loss and the testing loss. (4) Comparative studies reveal the ceiling of a DNN's complexity.

\section{Related Work}

\label{sec:related work}
We limit our discussion within the literature of understanding representations of DNNs. In general, previous studies can be roughly summarized into the following two types.

$\bullet$ The first type is the semantic explanations for DNNs.
Some studies directly visualized knowledge representations encoded in the DNN~\cite{zeiler2014visualizing,mahendran2015understanding,yosinski2015understanding,dosovitskiy2016inverting,simonyan2017deep}. 
Other methods estimated the pixel-wise attribution to the network output~\cite{zhou2014object,selvaraju2017grad,fong2017interpretable,kindermans2017learning,zhou2016learning}. The LIME~\cite{ribeiro2016should} and SHAP~\cite{lundberg2017unified} extracted important input units that directly contributed to the output. Some visual explanations reveal certain insightful properties of DNNs. \citet{fong2017interpretable} analyzed how multiple filters jointly represented a certain semantic concept. 
In contrast, in this paper, we propose to investigate the representation capacity from the perspective of transformation complexity encoded in DNNs.

$\bullet$ The second type is to analyze the feature representation capacity of DNNs. The stiffness~\cite{fort2019stiffness} was proposed to diagnose the generalization capacity of DNNs. \citet{xu2018understanding} applied the Fourier analysis to explain the generalization capacity of DNNs. The CLEVER score~\cite{weng2018evaluating} was used to estimate the robustness of DNNs. \citet{wolchover2017new,shwartz2017opening} proposed the information bottleneck theory and used mutual information to quantify the information encoded in DNNs. \citet{xu2017information,achille2018information,goldfeld2019estimating} further extended the information bottleneck theory to constrain the feature representation to learn more disentangled features. \citet{chen2018learning} selected instance-wise features based on mutual information to interpret DNNs. \citet{kornblith2019similarity} used the canonical correlation analysis to compare features from the perspective of similarity.

Unlike previous studies, we focus on the complexity of transformations encoded in DNNs. Previous methods on the complexity of DNNs can be summarized as follows:

$\bullet$ Computational complexity and difficulty of learning a DNN: \citet{blum1989training} proved that learning a one-layer network with a sign activation function was NP-hard. 
\citet{livni2014computational} further discussed the computational complexity when the DNN used other activation functions.
\citet{boob2018complexity,manurangsi2018computational} proved learning a two-layer ReLU network was also NP-hard.
\citet{arora2016understanding} showed that it was possible to learn a ReLU network with one hidden layer in polynomial time, when the dimension of input was constant.

$\bullet$ Architectural complexity and representation complexity:
\citet{raghu2017expressive} proved that the maximal complexity of features grew exponentially along with the increase of the DNN depth. \citet{pascanu2013construct,zhang2016architectural} proposed three metrics to measure the architectural complexity of recurrent neural networks. 
To estimate the maximal representation capacity, \cite{liang2017fisher,cortes2017adanet} applied the Rademacher complexity.
\cite{NIPS2019_8609} analyzed the complexity of features in a DNN by comparing them with features learned by a linear classifier.

Unlike analyzing the complexity of the DNN based on its architecture, we aim to measure the complexity of feature transformations learned by the DNN. We define three types of transformation complexity, which provide new perspectives to understand the DNN.

\begin{figure*}[t]
	\centering
	\includegraphics[width=\linewidth]{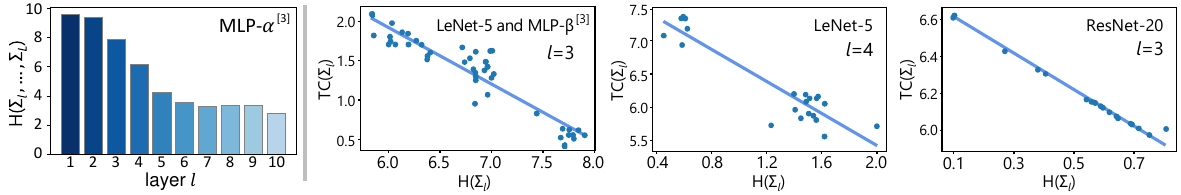}
	\vspace{-20pt}
	\caption{(left) The complexity of transforming feature {$T_{l}$} to the output, \emph{i.e.,} {$H(\Sigma_{l},\dots,\Sigma_{L})$}. The complexity decreases as we use features of higher layers. (right) The negative correlation between the transformation complexity {$H(\Sigma_{l})$} and the entanglement {$TC(\Sigma_{l})$} of DNNs with similar activation rates.}
	\label{fig:decrease_neg_correlation}
\end{figure*}

\section{Transformation Complexity}
\label{sec:tranformation complexity}

\textbf{Strong connection between gating states and the transformation complexity.}
Given the input {$x\in X$} and the target label {$y\in Y$}, the DNN is learned to map {$x$} to {$y$}. Layerwise transformations of mapping {$x$} to {$y$} can be roughly represented by
\begin{equation}
\begin{aligned}
    &y=g(z),\\
    &z=W_{L+1} \dots \boldsymbol{\sigma}_2(W_2 \boldsymbol{\sigma}_1(W_1x+b_1)+b_2)\dots +b_{L+1}
\end{aligned}
\label{eq:layerwise}
\end{equation}
where {$g$} denotes the optional layer on the top, \emph{e.g.,} the softmax layer.
{$z$} denotes the output feature before the top layer. \textbf{The network module {$z=W_{L+1} \dots \boldsymbol{\sigma}_2(W_2 \boldsymbol{\sigma}_1(W_1x+b_1)+b_2)\dots +b_{L+1}$} is a piecewise linear model.}

Specifically, {$W_{l}$} and {$b_{l}$} denote the weight and the bias of the {$l$}-th linear layer.
Let {$\sigma_l = [\sigma_{l,1},\sigma_{l,2},\dots,\sigma_{l,D}]\in\{0,1\}^D$} denote gating states of the {$l$}-th gating layer.
{$\boldsymbol{\sigma}_{l}=diag(\sigma_{l,1},\sigma_{l,2},\dots,\sigma_{l,D})$} is a diagonal matrix with {$(\sigma_{l,1},\sigma_{l,2},\dots,\sigma_{l,D})$} as its main diagonal. Gating layers include the ReLU, max-Pooling, and Dropout layer. Take the ReLU layer as an example\footnote{Please see Appendix \ref{sec:app_gating_layers} for more details about other types of gating layers.}. If the {$d$}-th dimension of the input feature is larger than 0, then we have {$\sigma_{l,d}=1$}; otherwise, {$\sigma_{l,d}=0$}.  Let {$\Sigma_{l}=\{\boldsymbol{\sigma}_{l}\}$} denote the random variable of gating states of the {$l$}-th gating layer. Given a certain input {$x$}, {$\boldsymbol{\sigma}=[\boldsymbol{\sigma}_1,\boldsymbol{\sigma}_2,\dots,\boldsymbol{\sigma}_{L}]$} represents concatenated and vectorized gating states of all gating layers. Accordingly, {$\Sigma=[\Sigma_1,\Sigma_2,\dots,\Sigma_{L}]$} denotes the set of gating states of all {$L$} gating layers over all samples\footnote{Please see Appendix \ref{sec:app_samples} for details about samples that are considered in implementations.\label{fn:samples}}.

Therefore, the layerwise transformation of mapping {$x$} to {$y$} can be represented as {$y=g(z)$} and {$z=\mathrm{\mathbf{W}} x+\mathrm{\mathbf{b}}$}, where {$\mathrm{\mathbf{W}}=W_{L+1}\boldsymbol{\sigma}_{L}W_L\cdots\boldsymbol{\sigma}_{2}W_2\boldsymbol{\sigma}_{1}W_1$} is the equivalent weight matrix, and {$\mathbf{b}$} is the equivalent bias term.
The piecewise linear module {$z=\mathrm{\mathbf{W}} x+\mathrm{\mathbf{b}}$} generates different gating states {$\boldsymbol{\sigma}_{1},\cdots,\boldsymbol{\sigma}_{L}$} for different inputs {$x$}, which lead to different values of {$\mathrm{\mathbf{W}}$} and {$\mathrm{\mathbf{b}}$}.
Therefore, we can roughly use the diversity of gating states to approximate the diversity of transformations. Thus, the entropy of gating states {$H(\Sigma)$} can represent the transformation complexity.

Note that not all small perturbations of the input {$x$} change the gating states, so gating states can be considered more directly related to the transformation complexity than the input.
In this way, we focus on gating layers in the DNN and define the following metrics to measure three types of complexities of transformations from {$x$} to {$y$}.

\textbf{Definition of the transformation complexity.}
In this section, we define three types of complexities of transformations in DNNs based on the information theory. 

$\bullet$ $H(\Sigma)$: the entropy of gating states among all inputs. {$H(\Sigma)$} measures the complexity of transformations that are encoded in gating layers. A larger {$H(\Sigma)$} indicates the DNN learns more complex transformations. The complexity {$H(\Sigma)$} can be decomposed as {$H(\Sigma)=H(\Sigma_1)+H(\Sigma_2|\Sigma_1)+\dots+H(\Sigma_{L}|\Sigma_1,\dots,\Sigma_{L-1})$}.

$\bullet$ $I(X;\Sigma)$: the complexity of transformations that are caused by the input. 
If the DNN does not use the random sampling operation or the dropout operation to introduce additional uncertainty that is not caused by the input in gating states {$\Sigma$}, then {$\Sigma$} is fully determined by {$X$}, \emph{i.e.,} {$H(\Sigma|X)=0$} and {$I(X;\Sigma)=H(\Sigma)-H(\Sigma|X)=H(\Sigma)$}.

$\bullet$ $I(X;\Sigma;Y)$: the complexity of transformations that are caused by inputs and used for inference, which is defined as {$I(X;\Sigma;Y)=I(X;Y)-I(X;Y|\Sigma)$}. {$I(X;Y|\Sigma)=H(X|\Sigma)-H(X|\Sigma,Y)$} measures the mutual information between {$X$} and {$Y$} that is irrelevant to gating layers.

Our definition of the complexity satisfies the following simple properties in DNNs, which ensures the trustworthiness of the complexity metric.

\textit{Property 1.} \textit{(Proof in Appendix \ref{sec:app_non-negative}) If the DNN does not introduce additional information that is not contained by the input $X$ (\emph{e.g.,} there are no operations of randomly sampling or dropout throughout the DNN), then we have $I(X;\Sigma;Y)\ge 0$.}

\textit{Property 2.}
\textit{If the DNN does not introduce additional complexity that is not caused by the input, then the complexity increases along with the number of gating layers.}

\textit{Property 3.}
\textit{If the DNN does not introduce additional complexity that is not caused by the input, then the complexity decreases when we use features of high layers for inference.
This property shows that the transformation complexity decreases through the layerwise propagation.}

Please see Appendix \ref{sec:app_increase} and \ref{sec:app_decrease} for formulations and proofs of Properties 2 and 3.

\textbf{Verification of the decrease of the complexity through layerwise propagation.}
Figure~\ref{fig:decrease_neg_correlation}(left) shows the change of the complexity of transforming the {$l$}-th layer feature {$T_{l}$} to the output, \emph{i.e.,} the entropy of gating states after the {$l$}-th layer {$H(\Sigma^\prime=[\Sigma_{l},\dots,\Sigma_{10}])$} encoded in the DNN during the training process on the MNIST dataset, which verified the decrease of the complexity through layerwise propagation. Experimental settings in Figure~\ref{fig:decrease_neg_correlation}(left) are introduced in Section~\ref{sec:analysis of DNNs}.

\textbf{The quantification of the transformation complexity.}
\label{sec:KDE}
There are three classic and widely-used non-parametric methods to estimate the entropy and mutual information of features, including the binning method~\cite{shwartz2017opening}, the ensemble dependency graph estimator (EDGE)~\cite{noshad2019scalable} and the kernel densitiy estimation (KDE)~\cite{kolchinsky2017estimating}.
Both the binning and the EDGE methods reduce the computational cost by discretizing the continuous features. However, the gating state $\Sigma$ is a discrete random variable, which cannot benefit from such methods.
Thus, we apply the KDE method to estimate the complexity {$H(\Sigma)$, $I(X;\Sigma)$} and {$I(X;\Sigma;Y)$}. 
Please see Appendix \ref{sec:app_KDE} for details about the KDE method of estimating {$H(\Sigma)$}, {$I(X;\Sigma)$} and {$I(X;\Sigma;Y)$}.
\textit{In Appendix \ref{sec:app_kappa}, we have also verified the high accuracy and the stability of using KDE to estimate {$H(\Sigma)$}, {$I(X;\Sigma)$} and {$I(X;\Sigma;Y)$}.}

\textbf{The difference between the transformation complexity and the information bottleneck.}
Note that the transformation complexity {$I(X;\Sigma)$} has essential difference from the {$I(X;Z)$} term in the information bottleneck theory~\cite{tishby2000information,wolchover2017new,shwartz2017opening}, where {$Z$} denotes the feature of an intermediate layer in DNNs.
The information bottleneck reflects the trade-off between {$I(X;Z)$} and {$I(Z;Y)$}, which leads to the approximate minimal sufficient statistics.
In the forward propagation, the feature {$Z$} contains all information encoded in the DNN, thereby forming a Markov process  {$X\!\to\! Z\!\to\! Y$}. Thus, given the feature {$Z$}, {$X$} and {$Y$} are conditional independent, \emph{i.e.,} {$I(X;Y|Z)=0$}.

However, the gating state {$\Sigma$} does not contain all information of the feature {$Z$}. 
Let us take the ReLU layer for an instance. In ReLU layers, the gating state $\Sigma$ only represents whether the elements in feature {$Z$} are positive. The information encoded in {$\Sigma$} cannot be directly used to infer {$Y$},
which makes the transformation complexity {$I(X;\Sigma)$} essentially different from {$I(X;Z)$} in mathematics.

\textbf{Advantages of investigating gating states over studying neural activations.}
{$I(X;\Sigma)$} based on gating states {$\Sigma$} is substantially more related to the transformation complexity than {$I(X;Z)$} based on neural activations {$Z$}.
Theoretically speaking, the entropy/mutual information of neural activations {$Z$} in an intermediate layer is affected by two aspects, \emph{i.e.,} the diversity of transformations and the continuous changes of input samples.
We have proven that gating states more directly reflect the transformation complexity in DNNs than neural activations {$z$}.

To be precise, any intermediate-layer neural activation {$z$} can be represented as a linear transformation on the input {$x$}, \emph{i.e.,} {$z=\mathrm{\mathbf{W}}x+\mathrm{\mathbf{b}}$}, where {$\mathrm{\mathbf{W}}$} and {$\mathrm{\mathbf{b}}$} are determined by gating states in nonlinear layers directly.
Therefore, gating states are the most direct factor that determines and reflects the diversity of linear subspaces in piecewise linear transformations, and are directly related to the representation power of a DNN. Whereas, the complexity of neural activations {$z$} are also affected by noisy signals in {$x$}, instead of purely being affected by the division of linear subspaces.

Therefore, in this study, we exclusively focus on the complexity of transformations {$(\mathrm{\mathbf{W}}, \mathrm{\mathbf{b}})$}, which can be considered as a more purified metric for a DNN's representation than the feature complexity.

\section{Analysis of DNNs Based on the Transformation Complexity}
\label{sec:analysis of DNNs}
\subsection{Strong correlation between the complexity and the disentanglement}
\label{sec:negative_correlation}

The disentanglement is a property of a DNN that measures the independence of different feature dimensions in the DNN. Stronger disentanglement sometimes leads to more interpretable features \cite{burgess2018understanding}, though the disentanglement is not necessarily equivalent to the discrimination power of features (which will be analyzed later).
Although intuitively, the disentanglement of gating states seems not related to the complexity, in this section, we prove a clear correlation between these two terms.

For gating states of the {$l$}-th gating layer {$\Sigma_{l}$}, the entanglement of transformations $TC(\Sigma_{l})$ measures the dependence between gating states $\sigma_{l,d}$ in different dimensions~\cite{achille2018emergence,ver2015maximally}.
Specifically, {$TC(\Sigma_{l})=KL(p(\sigma_{l})\Vert {\prod}_d p(\sigma_{l,d}))$}, where {$p(\sigma_{l,d})$} denotes the marginal distribution of the state of the $d$-th gate in {$\sigma_{l}$}. 
Let us assume that {$p(\sigma_{l,d})\sim \textrm{Bernoulli}(a_{l,d})$}, where {$a_{l,d}$} is the activation rate of the $d$-th dimension in the $l$-th gating layer over all samples.
In particular, {$TC(\Sigma_{l})$} is zero if and only if all dimensions of {$\sigma_{l}$} are independent with each other.
In this case, we say all dimensions in {$\Sigma_{l}$} are disentangled.

For DNNs with similar activation rates, we prove the negative correlation between the complexity {$H(\Sigma_{l})$} and the entanglement {$TC(\Sigma_{l})$} (proof in Appendix \ref{sec:app_negative_correlation}).
\begin{equation}
	H(\Sigma_{l})+TC(\Sigma_{l})=C_{l},\quad
	C_{l} = {\sum}_{d}H_{l,d},
	\label{eq:negative_HS}
\end{equation}
where {$H_{l,d}=-a_{l,d}\log a_{l,d}-(1-a_{l,d})\log(1-a_{l,d})$}. We can roughly consider {$C_l$} as a constant if the average activation rate over all dimensions {$\{a_{l,d} | 1\leq d\leq D\}$} in the {$l$}-th layer converges to a certain number.
Furthermore, we extend conclusions in Eq.~(\ref{eq:negative_HS}) and prove that the negative correlation between the complexity and the entanglement of transformations in the DNN still holds true for {$I(X;\Sigma_{l})$} and {$I(X;\Sigma_{l};Y)$}.

\begin{equation}
	\begin{aligned}
	 I(X;\Sigma_{l})&+TC(\Sigma_{l}) = C_{l}-H(\Sigma_{l}|X)\\
	I(X;\Sigma_{l};Y)&+\!\!\underbrace{(TC(\Sigma_{l})-TC(\Sigma_{l}|Y))}_{\textrm{multi-variate mutual information used to infer }Y}\\
	&= C_{l}-C_{l|Y}-(H(\Sigma_{l}|X)-H(\Sigma_{l}|X,Y))
	\end{aligned}
\end{equation}

\textbf{Whether the entanglement of features is good for classification or not?}
As is shown above, given a fixed activation rate in a DNN, the decrease of complexity leads to the increase of entanglement.
The entangled representations (or gating states) are supposed to occur in the following two cases.
One case is good for the classification, but the other is not.

(1) The first case is when the DNN fails to extract meaningful patterns from the input.
Many pixels in an input image are actually entangled with each other to represent a certain concept.
The successful extraction of features for meaningful concepts is to summarize all information of such pixel-wise entanglement into a certain intermediate-layer filter, which makes the intermediate-layer feature more disentangled.
The lack of the transformation complexity of a DNN usually will lead to high entanglement and hurt the classification performance.

(2) The second case is when the DNN has successfully selected a few discriminative and reliable features from input images in high convolutional layers for classification.
Sometimes, the number of reliable features is much less than the filter number.
In other words, there exists a significant redundancy of feature representations, \emph{i.e.,} multiple filters may represent similar features in high convolutional layers, which leads to high entanglement.
Such an entanglement (redundancy) of reliable features is good for classification.
More intuitive examples for these two cases are shown in Appendix \ref{sec:entanglement_example}.

\textbf{Verification of the strong correlation between the complexity and the disentanglement (Eq.~(\ref{eq:negative_HS})).}
We learned 21 LeNet-5 models and 21 MLP-$\beta$\footnote{{Please see the paragraph \textit{Experimental settings}, Section~\ref{sec:analysis of DNNs} for network architectures and experimental settings.}\label{fn:exp_setting}} models with different initialized parameters on the MNIST dataset. These models shared similar activation rates on each dimension in the $l$-th gating layer (we used {$l=3,4$}). Thus, we quantified {$H(\Sigma_{l})$} and {$TC(\Sigma_{l})$} (Please see Appendix \ref{sec:app_KDE} for the quantification of {$TC(\Sigma_l)$}) over the 42 models. We also conducted such an experiment on 21 ResNet-20 (RN-20) models with $l=3$.
Figure~\ref{fig:decrease_neg_correlation}(right) shows the negative correlation between {$H(\Sigma_{l})$} and {$TC(\Sigma_{l})$}, which was verified using different layers of DNNs with different architectures.

\begin{figure}[t]
	\centering
	\includegraphics[width=\linewidth]{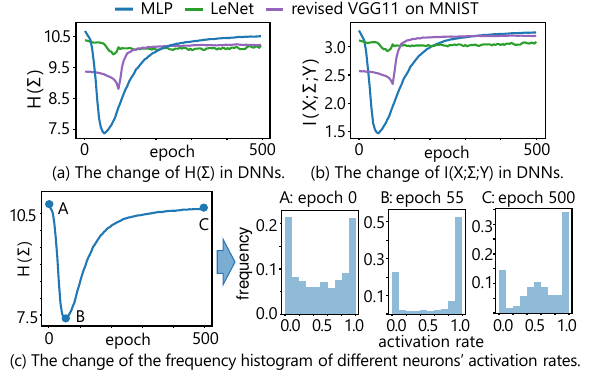}
	\vspace{-15pt}
	\caption{The change of the transformation complexity and the activation rates in traditional stacked DNNs. In most cases, both $H(\Sigma)$ and $I(X;\Sigma;Y)$ decreased first, and increased later.}
	\label{fig:change_tra}
\end{figure}

\subsection{Comparative studies to diagnose the representation capacity of DNNs}
\label{sec:exp}

\begin{figure}[t]
	\centering
	\includegraphics[width=\linewidth]{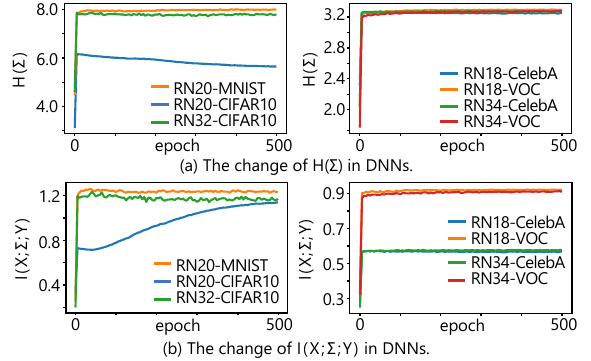}
	\vspace{-20pt}
	\caption{The change of the transformation complexity in residual networks. Both $H(\Sigma)$ and $I(X;\Sigma;Y)$ increased monotonously during the training process in most residual networks.}
	\label{fig:change_res}
	\vspace{-5pt}
\end{figure}

\textbf{The change of the transformation complexity during the learning of DNNs.}
Figure~\ref{fig:change_tra} and Figure~\ref{fig:change_res} shows the change of three types of transformation complexities encoded in DNNs during the training process. Note that {$H(\Sigma)=I(X;\Sigma)$}. We found three phenomena in the change of complexity:

\textit{Phenomenon 1.} For most traditional stacked DNNs (like MLPs\footref{fn:exp_setting} and LeNets), both {$H(\Sigma)$} and {$I(X;\Sigma;Y)$} decreased first, and increased later. Figure~\ref{fig:change_tra}(c) shows the frequency histogram of different neurons’ activation rates of gating states in Epoch 0, 55, and 500. In Epoch 0, gating states were usually randomly activated, which led to a large value of {$H(\Sigma)$}. The learning process gradually removed noisy activations, which reduced {$H(\Sigma)$} and achieved the minimum complexity in Epoch 55. Then, the DNN mainly learned complex transformations to boost the performance, which made {$H(\Sigma)$} begin to increase. This indicated that DNNs dropped noisy features in early stages of the training process, then learned useful information for the inference. 
In fact, the drop of noisy features in very early iterations has also been supported by observations in other studies~\cite{liu2021trap}. In very early iterations, initial weights irrelevant to the task were usually removed, which reduces the transformation complexity. Besides, \citet{liu2021trap} also showed that the decrease of the model diversity in very early iterations exactly aligned with the first stage in the epoch-wise double descent.

\textit{Phenomenon 2.} For residual DNNs with skip-connections and a few traditional DNNs (like the VGG-16 trained on the Pascal VOC dataset; see Appendix \ref{sec:vgg_complexity}), the complexity increased monotonously during the early stage of the training process, and saturated later. This indicated that noisy features had little effect on DNNs with skip-connections in early stages of the learning process, which implied the temporal double-descent phenomenon~\cite{nakkiran2019deep,heckel2020early}.

\begin{figure}[t]
	\centering
	\includegraphics[width=\linewidth]{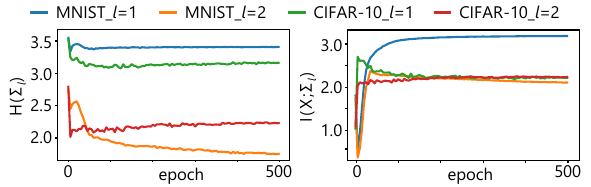}
	\vspace{-20pt}
	\caption{The change of {$H(\Sigma_{l})$} and {$I(X;\Sigma_{l})$} in VAEs learned on the MNIST dataset and the CIFAR-10 dataset. The difference between {$H(\Sigma_{l})$} and {$I(X;\Sigma_{l})$} gradually decreased during the training process.}
	\label{fig:vae}
\end{figure}

\textit{Phenomenon 3.} In particular, let us focus on DNNs, which introduce additional uncertainty that is not caused by the input.
As introduced in Section~\ref{sec:introduction}, a typical case is the VAE~\cite{kingma2013auto}. VAEs use randomly sampling of the latent code, and make {$H(\Sigma)\ne I(X,\Sigma)$}. 
Thus, in this experiment, we studied the change of {$H(\Sigma)$} and {$I(X;\Sigma)$} in VAEs.
We were given a VAE\footref{fn:exp_setting}, in which both the encoder and the decoder had two FC layers. We added a classifier with two FC layers and two ReLU layers after the encoder. The VAE was trained on the MNIST dataset and the CIFAR-10 dataset, respectively. Figure~\ref{fig:vae} shows the complexity of transformations encoded in each gating layer of the classifier. The difference between {$H(\Sigma_{l})$} and {$I(X;\Sigma_{l})$} gradually decreased during the training process. This indicated that the impact of inputs on the transformation complexity kept increasing. At the same time, the noisy features encoded in the DNN kept decreasing.

\textbf{Maximum complexity: exploring the ceiling of a DNN's  complexity.}
In order to examine whether a DNN always encoded more complex transformations when it dealt with more complex tasks, we first constructed a set of different tasks with various complexities.
Specifically, we designed a set of knowledge-distillation tasks with different complexity levels. In other words, we defined different \textit{task MLPs} with various complexities and distilled the knowledge of the \textit{task MLP} to a \textit{target MLP}.
We explored whether a target MLP would learn more complex transformations, when it distilled knowledge from a more complex task MLP.

Each of the task MLPs was assigned with randomly initialized parameters without further training. 
A task MLP consisted of {$n$} ReLU layers and FC layers with the width of 1024.
The task MLP took gray-scale CIFAR-10 images as input, and generated a 1024-dimensional vector as output.
We learned the target MLP to reconstruct this output vector with an MSE loss.

Since the task MLP was randomly parameterized, the task complexity would be high, when the number of ReLU layers {$n$} in the task MLP was large.
Therefore, we considered that the complexity of 
distilling knowledge of this task MLP into the target MLP was at the {$n$}-th level.
We conducted multiple experiments to train different target MLPs.
Each target MLP had a specific depth with 6, 12, 18 or 24 ReLU layers and FC layers, and each layer had 1024 neurons.
We trained these target MLPs to regress task MLPs of different complexities, {$n=0,1,\dots,31$}.

\begin{figure}[t]
	\centering
	\includegraphics[width=\linewidth]{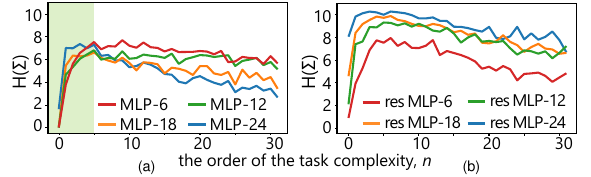}
	\vspace{-20pt}
	\caption{The change of target MLPs' transformation complexity along with the increase of task complexity $n$. We used MLPs with (a) traditional stacked architectures and  (b) residual architectures.}
	\label{fig:task-compl}
\end{figure}

\textit{Findings from stacked networks.} Figure~\ref{fig:task-compl}(a) compares the complexity of transformations encoded in target MLPs (with the traditional stacked architecture) for different tasks.\\
\textbf{(1)} For the task of low complexity, deep MLPs learned more complex transformations than shallow MLPs.\\
\textbf{(2)} However, as the complexity of the task increased, the complexity of transformations encoded in shallow MLPs was usually higher than that encoded in deep MLPs.\\
\textit{In other words, the transformation complexity did not monotonously increase along with the depth of the DNN.}
This phenomenon shows the ceiling of a DNN's complexity.

\textit{Findings from  residual networks.} Besides above target MLPs, we further designed new target MLPs with skip-connections, which were termed residual MLPs. We added a skip-connection to each FC layer in each of above target MLPs. The complexity of transformations encoded in residual MLPs was shown in Figure~\ref{fig:task-compl}(b).\\
\textbf{(1)} Deep residual MLPs always encoded higher transformation complexity than shallow residual MLPs.\\
\textbf{(2)} We also found that when we gradually increased the task complexity to train target MLPs, the transformation complexity encoded in target MLPs increased along with the increase of the task complexity in the beginning.\\
\textbf{(3)} However, when the task complexity exceeded a certain limit, the transformation complexity saturated and started to decrease.\\
\emph{I.e.,} \textit{the transformation complexity did not keep increasing when the DNN was forced to handle more and more complex tasks.}
This phenomenon indicated the ceiling of a DNN's complexity from another perspective.

\subsection{Learning  a DNN with the minimum complexity}
\label{sec:learning}

\textbf{Minimum complexity.}
A DNN may use over-complex transformations for prediction, \emph{i.e.,} the complexity {$I(X;\Sigma;Y)$} does not always represent the real complexity of the task. In this section, we develop a method to avoid learning an over-complex DNN. The basic idea is to use the following loss to quantify and penalize the complexity of transformations during the training process.
\begin{equation}
\begin{aligned}
    \mathcal{L}=&\mathcal{L}_{\textrm{task}}+\lambda\mathcal{L}_{\textrm{complexity}},\\
	\mathcal{L}_{\textrm{complexity}}=&{\sum}_{l=1}^L \! H(\Sigma_{l})\!=\!{\sum}_{l=1}^L \{-\mathbb{E}_{\sigma_{l}}[\log p(\sigma_{l})]\}
	\label{eqn:compl-loss}
\end{aligned}
\end{equation}
The first term {$\mathcal{L}_{\textrm{task}}$} denotes the task loss, \emph{e.g.,} the cross-entropy loss for object classification.
The second term {$\mathcal{L}_{\textrm{complexity}}$} penalizes the complexity of transformations encoded in the DNN. $\lambda$ is a positive scalar.
To simplify the computation of {$p(\sigma_{l})$}, we use an energy-based model (EBM)~\cite{lecun2006tutorial,gao2018learning} with parameters {$\theta_f$}.
The EBM is a widely-used method to model a high-dimensional distribution~\cite{du2019implicit,pang2020learning}, \emph{e.g.,} {$p_{\theta_f}(\sigma_l)$} in this paper.
\begin{equation}
	p_{\theta_f}(\sigma_{l})=\frac{1}{Z(\theta_f)}\exp [f(\sigma_{l};\theta_f)]\cdot q(\sigma_{l}),
\end{equation}
where {$q(\sigma_{l})$} is a prior distribution defined as {$q(\sigma_{l})=\prod_d q(\sigma_{l,d})$} and {$q(\sigma_{l,d})\!\!\sim\!\! \textrm{Bernoulli}(a_{l,d})$}.
{$f(\sigma_{l};\theta_f)\!\in\!\mathbb{R}$} is implemented as the scalar output of a ConvNet with parameters {$\theta_f$} and on the input {$\sigma_l$}~\cite{gao2018learning}.
The constant {$Z(\theta_f)=\int_{\sigma_{l}}q(\sigma_{l})\exp [f(\sigma_{l};\theta_f)] \mathrm{d}\sigma_{l}$} is for normalization.

The EBM is learned via the maximum likelihood estimation (MLE), {$\hat{\theta}_f\!=
\!\mathop{\arg\!\max}_{\theta_f}\mathbb{E}_x [\log p_{\theta_f}(\sigma_{l})_{\text{given }x}]$}, where {$\sigma_{l}$} is the vector of gating states in the $l$-th gating layer for input sample $x$. We follow~\cite{gao2018learning} to optimize the EBM parameters {$\theta_f$} with Markov Chain Monte Carlo.
Note that the gating state {$\sigma_{l}$} is not differentiable \emph{w.r.t.} the network parameters.
To this end, the ReLU operation can be approximated using the Swish function {$\textrm{ReLU}(x)\!\!=\!\!x\!\odot\!\sigma_{l}\!\!=\!\!x\!\odot\! \textrm{sigmoid}(\beta x)$}~\cite{ramachandran2017searching}, where { $\odot$} denotes the element-wise multiplication.
This enables us to use {$\mathcal{L}_{\textrm{complexity}}$} to learn network parameters. 
During the training process, the EBM and the original DNN are trained alternatively.
We discuss the computational cost of training DNNs with the complexity loss in Appendix \ref{subsec:app_computational_cost}.

\begin{figure}[t]
	\centering
	\includegraphics[width=\linewidth]{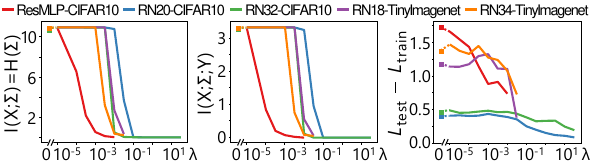}
	\vspace{-2em}
	\caption{The complexity of transformations and the gap between the training loss and the testing loss of the learned minimum-complexity DNNs. The left-most point in each subfigure at $\lambda=0$ refers to DNNs learned by only using the task loss.}
	\label{fig:minimum}
	\vspace{-5pt}
\end{figure}

\begin{figure}[t]
	\centering
	\includegraphics[width=\linewidth]{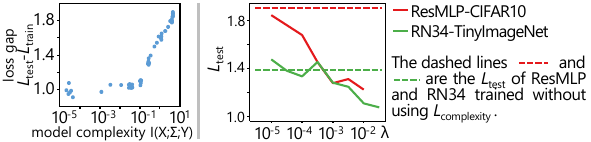}
	\vspace{-2em}
	\caption{(left) The positive correlation between transformation complexity and the loss gap. (right) Decrease of the testing loss along with the increase of the weight of the complexity loss.}
	\label{fig:rel-loss}
\end{figure}

\textbf{Validation of the utility of the complexity loss.}
The complexity loss reduced the transformation complexity and the gap between the training loss and the testing loss.
We added the complexity loss to the last four gating layers in each DNN to train the residual MLP\footnote{The residual MLP had the similar architecture to the one in the \textit{Findings from residual networks} paragraph in Section~\ref{sec:exp}, with 10 FC layers and 9 ReLU layers. Both inputs and features were 3072-d vectors.}, ResNet-20/32~\cite{he2016deep} (RN-20/32) on the CIFAR-10 dataset, and to train ResNet-18/34 (RN-18/34) on the first ten classes in the Tiny ImageNet dataset.  We repeatedly trained these DNNs with different values of $\lambda$. In particular, when $\lambda=0$, DNNs were learned only with $\mathcal{L}_{\textrm{task}}$, which can be taken as baselines.

\begin{table}[t]
    \centering
    \caption{Transformation complexity $H(\Sigma)$ in each layer of normally trained DNNs (termed \textit{normal}) and adversarially trained DNNs (termed \textit{AT}), which were trained on the CIFAR-10 dataset. Adversarially trained DNNs usually exhibited lower transformation complexity.
    \vspace{3pt}}
    \label{tab:adv-complexity}
    \resizebox{\linewidth}{!}{
    \begin{tabular}{c|c c| c c| c c| c c}
        \hline
        \multirow{2}{*}{Model} & \multicolumn{2}{c|}{RN-20} & \multicolumn{2}{c|}{RN-32} & \multicolumn{2}{c|}{RN-44} & \multicolumn{2}{c}{LeNet} \\
         & Normal  & AT  &  Normal & AT &  Normal  & AT  &  Normal & AT \\
        \hline
        Layer 1 & \textbf{3.845} & 2.979 &\textbf{2.718} & 2.507 & \textbf{3.938} & 1.600 & \textbf{7.624} & 5.358 \\ 
        Layer 2 & \textbf{6.079} &  4.485 &\textbf{5.660} & 4.370  & \textbf{5.426}  & 3.374 & \textbf{7.417} & 1.216\\
        Layer 3 & \textbf{6.671 } & 6.573 &\textbf{ 6.817} & 6.786 & 6.395 & \textbf{6.828}  &\textbf{ 10.966} & 10.949 \\
        \hline
    \end{tabular}
    }
\end{table}

Figure~\ref{fig:minimum} shows the complexity and the gap between the training loss and the testing loss of DNNs learned with different $\lambda$ values. We found that $H(\Sigma)$ usually decreased along with the increase of $\lambda$. $I(X;\Sigma;Y)$ also decreased along with the increase of $\lambda$, which verified that the complexity effectively reduced
the model's complexity.
We also found that the decrease of transformation complexity reduced the gap between the testing loss and the training loss.

Figure~\ref{fig:rel-loss}(left) demonstrated the positive correlation between the transformation complexity and the gap between the training loss and the testing loss. 
We assigned different weights $\lambda$ of the complexity loss to learn residual MLPs with different transformation complexities, using the CIFAR-10 dataset. We found that when the transformation complexity was high, the positive correlation was significant. This positive correlation further validated the effectiveness of the complexity loss on reducing the gap between the training loss and the testing loss.
Moreover, Figure~\ref{fig:rel-loss}(right) shows that the testing loss dropped significantly when we increased the weight $\lambda$ of the complexity loss.
Appendix \ref{subsec:app_compare_l1_l2} also shows that the complexity loss was superior to traditional {$L_1$} and {$L_2$} regularization methods, in terms of maintaining the classification accuracy and decreasing the model complexity.

Note that during the training process with the complexity loss, the activation rate {$a_{l,d}$} of the {$l$}-th gating layer would change ({$C_l$} in Eq. (\ref{eq:negative_HS}) may change), thus the entanglement of gating states {$TC(\Sigma_l)$} would not necessarily increase along with the decrease of the complexity.
The more entangled representations usually indicate more redundant features.

\begin{figure}[t]
    \centering
    \includegraphics[width=\linewidth]{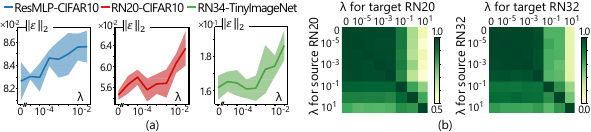}
    \vspace{-2.3em}
    \caption{(a) Increase of the minimum $L_2$ norm of the adversarial perturbations along with the increase of the weight of the complexity loss {$\lambda$}. (b) Adversarial transferability between DNNs learned with different weights of the complexity loss {$\lambda$}.}
    \label{fig:robustness-transferability}
    \vspace{-10pt}
\end{figure}

\textbf{The transformation complexity had a close relationship with adversarial robustness, adversarial transferability, and knowledge consistency.}
We set different values of the weight $\lambda$ in Eq. (\ref{eqn:compl-loss}) to learn DNNs with different transformation complexities.
For each $\lambda$ value, we repeatedly trained six residual MLPs on the CIFAR-10 dataset, six RN-20/32's on the CIFAR-10 dataset, and six RN-34's on the Tiny ImageNet dataset, with different random initializations.
We followed~\cite{wang2020unified} to measure the minimum $L_2$ norm of adversarial perturbations computed \emph{w.r.t.} a certain attacking utility to quantify the adversarial robustness of DNNs.
Figure \ref{fig:robustness-transferability}(a) shows the minimum $L_2$ norm of adversarial perturbations towards each DNN.
We found that DNNs with low transformation complexity usually exhibited high adversarial robustness, while DNNs with high transformation complexity were usually sensitive to adversarial perturbations.
Besides, Table~\ref{tab:adv-complexity} compares the transformation complexity in normally trained DNNs and adversarially trained DNNs.
Results show that adversarially trained DNNs, which were more robust than normally trained DNNs, usually exhibited lower complexity.
This also verifies the negative correlation between transformation complexity and adversarial robustness.

Following settings in~\cite{wang2020unified}, we also measured the adversarial transferability between DNNs with different transformation complexities.
Figure \ref{fig:robustness-transferability}(b) shows the adversarial transferability between ResNet-20/32's with different transformation complexities.
We found that adversarial perturbations for complex DNNs could not be well transferred to simple DNNs. However, adversarial perturbations for simple DNNs could be transferred to complex DNNs.
This reflected that simple DNNs encoded common knowledge that could be transferred to DNNs learned for the same task.

Furthermore, we followed \cite{liang2019knowledge} to explore the knowledge consistency between DNNs with different transformation complexities.
We found that each pair of simple DNNs usually encoded similar knowledge representations (exhibiting high knowledge consistency), while complex DNNs are more likely to encode diverse knowledge.
This demonstrated the reliability of features learned by simple DNNs.
Please refer to Appendix \ref{subsec:app_detail_robust_trans_consis} for experimental results on knowledge consistency.

\textbf{Experimental settings.}
We conducted a set of comparative studies on the task of classification using the MNIST~\cite{lecun1998gradient}, CIFAR-10~\cite{krizhevsky2009learning}, CelebA~\cite{liu2015deep}, Pascal VOC 2012~\cite{everingham2015pascal}, and Tiny ImageNet~\cite{tiny_imagenet} datasets. For the MNIST dataset and the CIFAR-10 dataset, we learned LeNet-5~\cite{lecun1998gradient}, the revised VGG-11~\cite{jastrzkebski2017three}, the pre-activation version of ResNet-20/32~\cite{he2016deep,he2016identity}, and the MLP. In particular, for the MNIST dataset, we learned three MLP models: \textit{MLP-MNIST} contained 5 fully connected (FC) layers with the width of 784-1024-256-128-64-10, \textit{MLP-$\alpha$} contained 11 FC layers with the width of 784-1024-1024-512-512-256-256-128-128-64-16-10, and \textit{MLP-$\beta$} contained 5 FC layers with the width of 784-1024-256-120-84-10\footnote{For comparison, we modified the architecture of MLP-MNIST and made the width of the last three FC layers be the same with the fully connected layers in the LeNet-5 network.}.
For the CIFAR-10 dataset, the architecture of the MLP was set as 3072-1024-256-128-64-10 (termed  \textit{MLP-CIFAR10}). 
For the CelebA, Pascal VOC 2012, and Tiny ImageNet datasets, we learned VGG-16~\cite{simonyan2017deep} and the pre-activation version of ResNet-18/34.
We used images cropped by bounding boxes for both training and testing. Please see Appendix \ref{sec:app_acc} for these DNNs' classification accuracy. We analyzed the transformation complexity of ReLU layers.

\section{Conclusion}
In this paper, we have proposed three complexity measures for feature transformations encoded in DNNs. We further prove the decrease of the transformation complexity through layerwise propagation. We also prove the strong correlation between the complexity and the disentanglement of transformations. Based on the proposed metrics, we develop a generic method to learn a minimum-complexity DNN, which also reduces the gap between the training loss and the testing loss, and influences adversarial robustness, adversarial transferability, and knowledge consistency. Comparative studies reveal the ceiling of a DNN's complexity. Furthermore, we summarize the change of the transformation complexity during the training process into two typical cases. As a generic tool, the transformation complexity enables us to understand DNNs from new perspectives.

\textbf{Acknowledgments.} This work is partially supported by National Key R\&D Program of China (2021ZD0111602), the National Nature Science Foundation of China (No. 61906120, U19B2043), Shanghai Natural Science Foundation (21JC1403800, 21ZR1434600), Shanghai Municipal Science and Technology Major Project (2021SHZDZX0102).
This work is also partially supported by Huawei Technologies Inc.

\bibliography{example_paper}
\bibliographystyle{icml2022}

%%%%%%%%%%%%%%%%%%%%%%%%%%%%%%%%%%%%%%%%%%%%%%%%%%%%%%%%%%%%%%%%%%%%%%%%%%%%%%%
%%%%%%%%%%%%%%%%%%%%%%%%%%%%%%%%%%%%%%%%%%%%%%%%%%%%%%%%%%%%%%%%%%%%%%%%%%%%%%%
% APPENDIX
%%%%%%%%%%%%%%%%%%%%%%%%%%%%%%%%%%%%%%%%%%%%%%%%%%%%%%%%%%%%%%%%%%%%%%%%%%%%%%%
%%%%%%%%%%%%%%%%%%%%%%%%%%%%%%%%%%%%%%%%%%%%%%%%%%%%%%%%%%%%%%%%%%%%%%%%%%%%%%%
\newpage
\appendix
\onecolumn

\section{Gating States in Gating Layers}
\label{sec:app_gating_layers}

In this section, we further discuss gating states of different gating layers, which is mentioned in Section \ref{sec:tranformation complexity} of the paper. Let $h_{l}\triangleq W_{l}(\boldsymbol{\sigma}_{l-1}(W_{l-1}\dots(W_2\boldsymbol{\sigma}_1(W_1x+b_1)+b_2)\dots+b_{l-1}))+b_{l}$ denote the input of the $l$-th gating layer. We consider the vectorized form of $h_{l}$. Given $h_{l}\in\mathbb{R}^D$, the formulation of $\boldsymbol{\sigma}_{l}$ in different gating layers is given as follows.

(1) ReLU layer. In this case, $\boldsymbol{\sigma}_{l}=diag(\sigma_{l}^1, \sigma_{l}^2, \dots, \sigma_{l}^D)\in\{0,1\}^D$, which is a diagonal matrix. If the $d$-th dimension of $h_{l}$ is larger than 0, then we have $\sigma_{l}^d=1$; otherwise, $\sigma_{l}^d=0$.

(2) Dropout layer. In this case, $\boldsymbol{\sigma}_{l}=diag(\sigma_{l}^1, \sigma_{l}^2, \dots, \sigma_{l}^D)\in\{0,1\}^D$, which is a diagonal matrix. If the $d$-th dimension of $h_{l}$ is not dropped, then we have $\sigma_{l}^d=1$; otherwise, $\sigma_{l}^d=0$.

(3) Max-Pooling layer. Since a pooling layer may change the size of the input, $\boldsymbol{\sigma}_{l}$ is not necessarily a square matrix. Let the output of the max-pooling layer be $\boldsymbol{\sigma}_{l} h_{l}\in\mathbb{R}^{D^\prime}$, \emph{i.e.} the input $h_{l}$ is divided into $D^\prime$ regions. In this case, we have $\boldsymbol{\sigma}_{l}\in\{0,1\}^{D^\prime\times D}$. If $(h_{l})_{d^\prime}$ is the largest element in the $d^\prime$-th region, then we have $(\boldsymbol{\sigma}_{l})_{d^\prime d}=1$; otherwise, $(\boldsymbol{\sigma}_{l})_{d^\prime d}=0$.

\section{Proofs of Important Conclusions}
\label{sec:app_proof}
This section proves the three properties mentioned in Section \ref{sec:tranformation complexity} of the paper, and also gives detailed proofs for other important conclusions in the paper.

\subsection{Non-negativity of the complexity $I(X;\Sigma;Y)$}
\label{sec:app_non-negative}
\textbf{Property 1.}\quad\textit{If the DNN does not introduce information besides the input $X$ (\emph{e.g.} there is no sampling operations or dropout operations throughout the DNN), we have $I(X;\Sigma;Y)\ge0$.}

\textit{\textbf{Proof.}} Recall that the mutual information is defined as
	\begin{equation}
		\begin{aligned}
		I(X;\Sigma;Y)=&I(X;Y)-I(X;Y|\Sigma)\\
		=&(H(Y)-H(Y|X))-(H(Y|\Sigma)-H(Y|X,\Sigma))\\
		=&(H(Y)-H(Y|\Sigma))-(H(Y|X)-H(Y|X,\Sigma))
		\end{aligned}
	\end{equation}
	If the DNN does not introduce additional information besides $X$, which means that $\Sigma$ is determined by $X$, then we have $H(Y|X)-H(Y|X,\Sigma)=I(\Sigma;Y|X)=0$. Therefore,
	\begin{equation}
		I(X;\Sigma;Y)=H(Y)-H(Y|\Sigma) \ge 0
		\label{equ:ISY}
	\end{equation}

\subsection{Increase of the complexity along with the increase of gating layers}
\label{sec:app_increase}

\textbf{Property 2.}\quad The complexity of transforming the input $x$ increases when we consider more gating layers.

\textbullet{ \textbf{$H(\Sigma_{1},\dots,\Sigma_{l}) \le H(\Sigma_{1},\dots,\Sigma_{l+1})$}

\textit{\textbf{Proof.}}
\begin{equation}
	\begin{aligned}
		&H(\Sigma_{1},\dots,\Sigma_{l}) - H(\Sigma_{1},\dots,\Sigma_{l+1})\\
		=&-H(\Sigma_{l+1}|\Sigma_{1},\dots,\Sigma_{l})\\
		=&\mathbb{E}_{\sigma_{1},\dots,\sigma_{l+1}}\left[\log p(\sigma_{l+1}|\sigma_{1},\dots,\sigma_{l})\right]\\
		\le& 0
	\end{aligned}
\end{equation}
}

\textbullet{ \textbf{$I(X;\{\Sigma_{1},\dots,\Sigma_{1}\}) \le I(X;\{\Sigma_{1},\dots,\Sigma_{l+1}\})$}
	
\textit{\textbf{Proof.}}
If the DNN does not introduce additional information besides the input during the forward propagation, then $\Sigma_{1}, \dots, \Sigma_{l}, \Sigma_{l+1}$ are all determined by $X$, thereby $H(\Sigma_{1},\dots,\Sigma_{l}|X)=H(\Sigma_{1},\dots,\Sigma_{l+1}|X)=0$. Therefore,
\begin{equation}
	\begin{aligned}
		&I(X;\{\Sigma_{1},\dots,\Sigma_{l}\}) - I(X;\{\Sigma_{1},\dots,\Sigma_{l+1}\})\\
		=& (H(\Sigma_{1},\dots,\Sigma_{l})-H(\Sigma_{1},\dots,\Sigma_{l}|X))-(H(\Sigma_{1},\dots,\Sigma_{l+1})-H(\Sigma_{1},\dots,\Sigma_{l+1}|X))\\
		=&H(\Sigma_{1},\dots,\Sigma_{l})-H(\Sigma_{1},\dots,\Sigma_{l+1})\\
		\le& 0
	\end{aligned}
	\label{eq:increase}
\end{equation}
}

\textbullet{ \textbf{$I(X;\{\Sigma_{1},\dots,\Sigma_{l}\};Y) \ge I(X;\{\Sigma_{1},\dots,\Sigma_{l+1}\};Y)$}
	
\textit{\textbf{Proof.}}
According to Eq.~(\ref{equ:ISY}), if there is no additional information besides the input throughout the DNN, then
\begin{equation}
	I(X;\{\Sigma_{1},\dots,\Sigma_{l}\};Y)=H(Y)-H(Y|\{\Sigma_{1},\dots,\Sigma_{l}\})
\end{equation} 
We can obtain the following inequality:
\begin{equation}
	\begin{aligned}
		& I(X;\{\Sigma_{1},\dots,\Sigma_{l}\};Y) - I(X;\{\Sigma_{1},\dots,\Sigma_{l+1}\};Y)\\
		=&(H(Y)-H(Y|\{\Sigma_{1},\dots,\Sigma_{l}\}))-(H(Y)-H(Y|\{\Sigma_{1},\dots,\Sigma_{l+1}\}))\\
		=&H(Y|\{\Sigma_{1},\dots,\Sigma_{l+1}\})-H(Y|\{\Sigma_{1},\dots,\Sigma_{l}\})\\
		=&-I(\Sigma_{l+1};Y|\{\Sigma_{1},\dots,\Sigma_{l}\})\\
		\le &0
	\end{aligned}
\end{equation}
}

\subsection{Decrease of the complexity through layerwise propagation}
\label{sec:app_decrease}
\textbf{Property 3.}\quad The complexity of transforming the intermediate-layer feature $t_{l}$ to the output $y$ decreases, when we use the feature of higher layers.

\textbullet{ \textbf{$H(\Sigma_{l},\dots,\Sigma_{L}) \ge H(\Sigma_{l+1},\dots,\Sigma_{L})$}
	
\textit{\textbf{Proof.}}
	\begin{equation}
		\begin{aligned}
		&H(\Sigma_{l},\dots,\Sigma_{L}) - H(\Sigma_{l+1},\dots,\Sigma_{L})\\
		=&H(\Sigma_{l}|\Sigma_{l+1},\dots,\Sigma_{L})\\
		=&-\mathbb{E}_{\sigma_{l},\dots,\sigma_{L}}\left[\log p(\sigma_{l}|\sigma_{l+1},\dots,\sigma_{L})\right]\\
		\ge& 0
		\end{aligned}
	\end{equation}
}

\textbullet{ \textbf{$I(T_{l-1};\{\Sigma_{l},\dots,\Sigma_{L}\}) \ge I(T_{l};\{\Sigma_{l+1},\dots,\Sigma_{L}\})$}
	
\textit{\textbf{Proof.}}
If the DNN does not introduce additional information besides the input during the forward propagation, then $\Sigma_{l}, \Sigma_{l+1}, \ldots, \Sigma_{L}$ are all determined by $T_{l-1}$, thereby $H(\Sigma_{l},\dots,\Sigma_{L}|T_{l-1})=0$. Therefore,
	\begin{equation}
		\begin{aligned}
		&I(T_{l-1};\{\Sigma_{l},\dots,\Sigma_{L}\}) - I(T_{l};\{\Sigma_{l+1},\dots,\Sigma_{L}\})\\
		=& (H(\Sigma_{l},\dots,\Sigma_{L})-H(\Sigma_{l},\dots,\Sigma_{L}|T_{l-1}))-(H(\Sigma_{l+1},\dots,\Sigma_{L})-H(\Sigma_{l+1},\dots,\Sigma_{L}|T_{l}))\\
		=&H(\Sigma_{l},\dots,\Sigma_{L})-H(\Sigma_{l+1},\dots,\Sigma_{L})\\
		\ge& 0
		\end{aligned}
		\label{eq:decrease}
	\end{equation}
}

\textbullet{ \textbf{$I(T_{l-1};\{\Sigma_{l},\dots,\Sigma_{L}\};Y) \ge  I(T_{l};\{\Sigma_{l+1},\dots,\Sigma_{L}\};Y)$}
	
\textit{\textbf{Proof.}}
According to Eq.~(\ref{equ:ISY}), if there is no additional information besides the input throughout the DNN, then
\begin{equation}
	I(T_{l-1};\{\Sigma_{l},\dots,\Sigma_{L}\};Y)=H(Y)-H(Y|\{\Sigma_{l},\dots,\Sigma_{L}\})
\end{equation} 
We can obtain the following inequality:
	\begin{equation}
		\begin{aligned}
		& I(T_{l-1};\{\Sigma_{l},\dots,\Sigma_{L}\};Y) - I(T_{l};\{\Sigma_{l+1},\dots,\Sigma_{L}\};Y)\\
		=&(H(Y)-H(Y|\{\Sigma_{l},\dots,\Sigma_{L}\}))-(H(Y)-H(Y|\{\Sigma_{l+1},\dots,\Sigma_{L}\}))\\
		=&H(Y|\{\Sigma_{l+1},\dots,\Sigma_{L}\})-H(Y|\{\Sigma_{l},\dots,\Sigma_{L}\})\\
		=&I(\Sigma_{l};Y|\{\Sigma_{l+1},\dots,\Sigma_{L}\})\\
		\ge &0
		\end{aligned}
	\end{equation}
}

\subsection{Strong correlations between the complexity  and the disentanglement of transformations}
\label{sec:app_negative_correlation}

Some previous studies used the entanglement (the multi-variate mutual information) to analyze the information encoded in DNNs. \cite{ver2015maximally} used $TC(X)$ to measure the correlation between different input samples. In comparison, in this paper, we apply $TC(\Sigma)$ to measure the independence between gating states of different dimensions. Intuitively, the disentanglement of gating states does not seem related to the complexity. Therefore, our contribution is to find out the strong correlation between the two factors which seem not related.

We consider the complexity of the transformation of a single gating layer, \emph{e.g.} $H(\Sigma_{l})$, $I(X;\Sigma_{l})$ and $I(X;\Sigma_{l};Y)$ for the $l$-th gating layer.

\textbullet{ \textbf{$H(\Sigma_{l})$}

\textit{\textbf{Proof.}} 
	\begin{equation}
	\begin{aligned}
	H(\Sigma_{l})+TC(\Sigma_{l}) =&H(\Sigma_{l})+KL(p(\sigma_{l})||\prod_d p(\sigma_{l}^{d}))\\
	=&\mathbb{E}_{\sigma_{l}}\left[\log \frac{1}{p(\sigma_{l})}\right]+\mathbb{E}_{\sigma_{l}}\left[\log \frac{p(\sigma_{l})}{\prod_{d}p(\sigma_{l}^{d})}\right]\\
	=&-\mathbb{E}_{\sigma_{l}}\left[\log\prod_dp(\sigma_{l}^{d})\right] ~\backslash\backslash~ p(\sigma_{l}^{d})\textrm{ does not depend on the input}\\
	=&C_{l}
	\end{aligned}
	\label{equ:negative_0}
	\end{equation}
Let us consider DNNs with similar activation rates $a_{l}^d$. Because $p(\sigma_{l}^{d})$ follows the Bernoulli distribution with the activation rate $a_{l}^d$, for DNNs with similar activation rates $a_{l}^d$, they share similar values of $C_{l}$. In this case, there is a negative correlation between $H(\Sigma_{l})$ and $TC(\Sigma_{l})$.
}

\textbullet{\textbf{$I(X;\Sigma_{l})$}

\textit{\textbf{Proof.}}
	\begin{equation}
		\begin{aligned}
		I(X;\Sigma_{l})+TC(\Sigma_{l}) 
		=& H(\Sigma_{l})-H(\Sigma_{l}|X)+KL(p(\sigma_{l})||\prod_d p(\sigma_{l}^{d}))\\
		=& C_{l} - H(\Sigma_{l}|X)
		\end{aligned}
	\end{equation}

If the DNN does not introduce additional information through the layerwise propagation, then the $X$ determines $\Sigma_{l}$, \emph{i.e.} $H(\Sigma_{l}|X)=0$. Thus, for DNNs with similar values of $C_{l}$, there is a negative correlation between $I(X;\Sigma_{l})$ and $TC(\Sigma_{l})$. 
}

\textbullet{\textbf{$I(X;\Sigma_{l};Y)$}
		
\textit{\textbf{Proof.}} According to the definition of $I(X;\Sigma_{l};Y)$, we have 
	\begin{equation}
		I(X;\Sigma_{l};Y) = I(X;\Sigma_{l})-I(X;\Sigma_{l}|Y)
	\end{equation}
	We have discussed the first term $I(X;\Sigma_{l})$ above, so we focus on the second term $I(X;\Sigma_{l}|Y)$, which measures the complexity of transformations that are unrelated to the inference. Similarly, the entanglement of the inference-irrelevant transformations is represented by
	\begin{equation}
		TC(\Sigma_{l}|Y)=\mathbb{E}_y (KL(p(\sigma_{l}|y)||\prod_d p(\sigma_{l}^{d}|y)))
	\end{equation}
	Then, we have
	\begin{equation}
		\begin{aligned}
		&I(X;\Sigma_{l}|Y)+TC(\Sigma_{l}|Y)\\
		=&H(\Sigma_{l}|Y)-H(\Sigma_{l}|X,Y)+TC(\Sigma_{l}|Y)\\
		=&\mathbb{E}_y\left[H(\Sigma_{l}|y)+KL(p(\sigma_{l}|y)||\prod_d p(\sigma_{l}^{d}|y))\right]-H(\Sigma_{l}|X,Y)\\
		=&\mathbb{E}_{\sigma_{l},y}\left[\log\frac{1}{p(\sigma_{l}|y)}+\log \frac{p(\sigma_{l}|y)}{\prod_d p(\sigma_{l}^{d}|y)} \right] - H(\Sigma_{l}|X,Y)\\
		=&-\mathbb{E}_{\sigma_{l},y}\left[\log \prod_d p(\sigma_{l}^{d}|y)\right]- H(\Sigma_{l}|X,Y)\\
		=&C_{l|Y}- H(\Sigma_{l}|X,Y)
		\end{aligned}
	\end{equation}
	If there is no additional information besides the input in the DNN, then  $H(\Sigma_{l}|X,Y)=0$. Thus, we have
	\begin{equation}
		\begin{aligned}
		&I(X;\Sigma_{l})+TC(\Sigma_{l})=C_{l}- H(\Sigma_{l}|X)=C_{l}\\
		&I(X;\Sigma_{l}|Y)+TC(\Sigma_{l}|Y)=C_{l|Y}-H(\Sigma_{l}|X,Y)=C_{l|Y}
		\end{aligned}
	\end{equation}
	
	Therefore, 
	\begin{equation}
		\begin{aligned}
		I(X;\Sigma_{l};Y) &= I(X;\Sigma_{l})-I(X;\Sigma_{l}|Y)\\
		&=(C_{l}-TC(\Sigma_{l}))-(C_{l|Y}-TC(\Sigma_{l}|Y))\\
		&=(C_{l}-C_{l|Y})-\underbrace{(TC(\Sigma_{l})-TC(\Sigma_{l}|Y))}_{\substack{\textrm{multi-variate mutual information}\\ \textrm{ used to infer $Y$}}}
		\end{aligned}
		\label{eq:IXSY_single}
	\end{equation}
	where the difference between $TC(\Sigma_{l})$ and $TC(\Sigma_{l}|Y)$ represents the entanglement of the transformations that are used to infer $Y$. For DNNs with similar activation rates, we can also roughly consider that these DNNs share similar values of $C_{l}$ and $C_{l|Y}$.  Thus, we can conclude that the higher complexity makes the DNN use more disentangled transformation for inference.
}

\section{About Values of $C_{l|Y}$}
\label{sec:app_value of C(l|Y)}
In experiments, we found that in most cases, for DNNs with similar activation rates $a_{l}^d$ in their corresponding layers, these DNNs usually shared similar values of $C_{l|Y}$. However, in some extreme cases, \emph{e.g.} when the DNN was learned from very few training samples, or when the target layer was very close to the input layer or the output layer, values of $C_{l|Y}$ of these DNNs were different from those values of other DNNs.

\section{Intuitive Examples for the Entanglement of Features}
\label{sec:entanglement_example}

This section introduces intuitive examples for the two cases of feature entanglement mentioned in Section \ref{sec:negative_correlation} of the paper.

For the first case, if we use an extremely simple DNN (\emph{e.g.} a two-layer neural network) to classify complex images, it is likely that the network fails to extract features for meaningful concepts.
The lack of representation power (transformation complexity) of a DNN usually leads to high entanglement of features and hurts the classification performance.
In other words, the intermediate-layer features in the DNN are still highly entangled, just like the entanglement of pixel colors in the image.

For the second case, if we use a sophisticated enough DNN (\emph{e.g.} a VGG-16) to classify very simple digits in the MNIST dataset. 
Then, features in high layers (\emph{e.g.} the {\small\texttt{conv5-3}} layer) are redundant to represent the simple knowledge in the digits, thus leading to entanglement of features.
In this case, due to the feature redundancy, features in some channels of the {\small\texttt{conv5-3}} layer may be similar to each other, which can be considered as feature entanglement.
Such an entanglement (redundancy) of reliable features is good for classification.

\section{Detailed Explanation of the KDE Method}
\label{sec:app_KDE}
The kernel density estimation (KDE) method was proposed by \cite{kolchinsky2017estimating}, and has been considered as a standard method to estimate the entropy and the mutual information. In this section, we briefly summarize key techniques of the KDE method, which are used to quantify the transformation complexity.

The KDE approach was proposed to estimate the mutual information between the input $X$ and the feature of an intermediate layer $T$ in a DNN~\cite{kolchinsky2017estimating,kolchinsky2019nonlinear}.
The KDE approach assumes that the intermediate-layer feature is distributed as a mixture of Gaussians. 
Since $T$ is a continuous variable, $H(T)$ can be negative.
The KDE method transforms each feature point into a local Gaussian distribution to approximate the accurate feature distribution.
Let $\hat{T}=T+\epsilon$ where $\epsilon\sim \mathcal{N}(0,\sigma_0^2I)$. Then, the distribution of $\hat{T}$ can be considered as a mixture of Gaussians, with a Gaussian centered on $T$. In this setting, previous studies~\cite{kolchinsky2017estimating,kolchinsky2019nonlinear,saxe2019information} shows that an upper bound for the mutual information with the input is
\begin{equation}
    I(\hat{T};X)\le -\frac{1}{P}\sum_i\log \frac{1}{P}\sum_j\exp\left(-\frac{1}{2}\frac{||t_i-t_j||^2}{\sigma_0^2}\right)
    \label{eq:quantify_IXT}
\end{equation}
where $P$ is the number of training samples, and $t_i$ denotes the intermediate-layer feature of the input sample $i$. Similarly, the upper bound for the mutual information \emph{w.r.t} the output $Y$ can be calculated as
\begin{equation}
\begin{aligned}
    I(\hat{T};Y)=&H(\hat{T})-H(\hat{T}|Y)\\
    =&-\frac{1}{P}\sum_i\log \frac{1}{P}\sum_j\exp\left(-\frac{1}{2}\frac{||t_i-t_j||^2}{\sigma_0^2}\right)\\
    &-\sum_{l=1}^L p_{l}\left[-\frac{1}{P}\sum_{i:Y_i=l}
    \log \frac{1}{P} \sum_{j:Y_j=l}\exp \left(-\frac{1}{2}\frac{||t_i-t_j||^2}{\sigma_0^2}\right)\right]
\end{aligned}
\end{equation}
where $L$ is the number of categories. $P_{l}$ denotes the number of samples belonging to the $l$-th category. $p_{l}=P_{l}/P$ denotes the probability of the category $l$.

\begin{figure}[h]
	\centering
	\begin{minipage}{0.28\linewidth}
		\centering
		\includegraphics[height=0.7\linewidth]{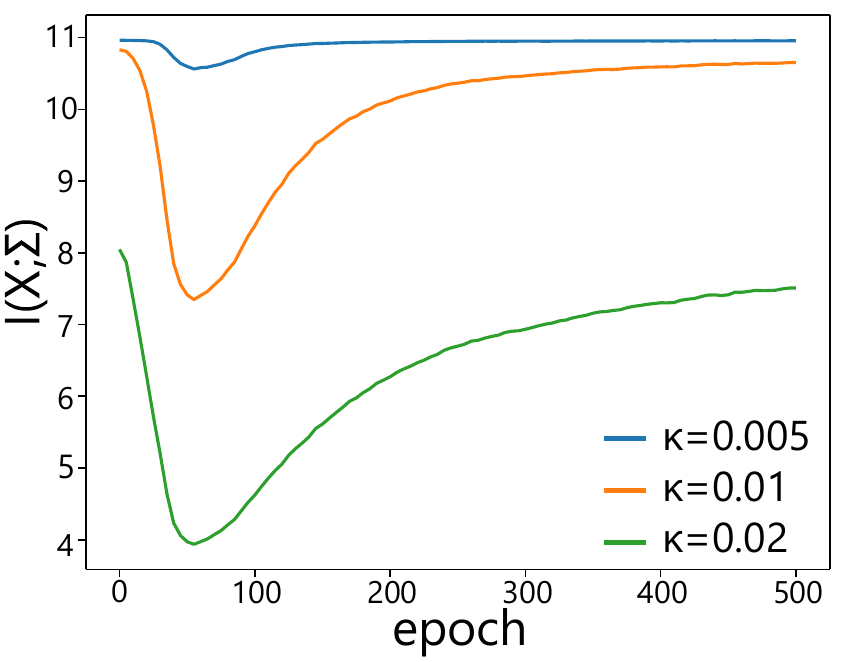}
	\end{minipage}
	\begin{minipage}{0.28\linewidth}
		\centering
		\includegraphics[height=0.7\linewidth]{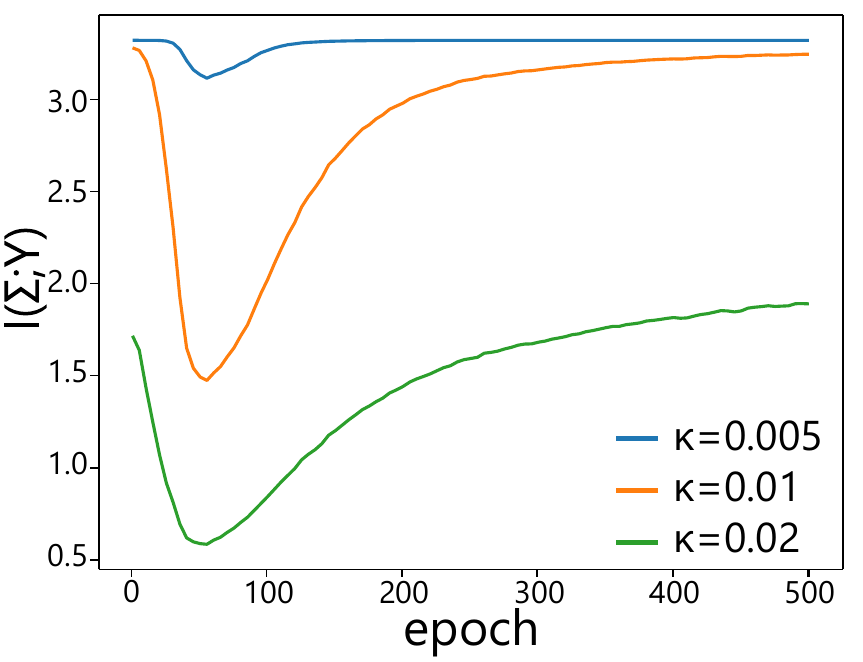}
	\end{minipage}
	\vspace{-5pt}
	\caption{The complexity calculated with different values of $\kappa$. 
	The $\kappa$ value only affected the scale of $I(X;\Sigma)$, but it did not affect the trend of the complexity change. Thus, given a fixed $\kappa$ value, the complexity of different DNNs can be fairly compared.}
	\label{fig:different_kappa}
\end{figure}

We use the KDE method to quantify the transformation complexity $H(\Sigma)$, $I(X;\Sigma)$ and $I(X;\Sigma;Y)$. The entropy of gating states $H(\Sigma_{l})$ is quantified as follows.
\begin{equation}
	H(\Sigma_{l})\le -\frac{1}{n}\sum_{j=1}^n\log \frac{1}{n}\sum_{k=1}^n
\exp\left(-\frac{1}{2}\frac{\Vert\sigma_{l,j}-\sigma_{l,k}\Vert_2^2}{\sigma_0^2}\right)
\label{eq:quantify_HS}
\end{equation}
where $n$ denotes the number of training samples. $\sigma_{l,j}$ and $\sigma_{l,k}$ denote the vectorized gating states of the $l$-th gating layer for the $j$-th sample and the $k$-th sample, respectively. $\sigma_0^2$ is quantified as $\sigma_0^2=\kappa\cdot Var(\Sigma_{l})$, where $Var(\Sigma_{l})=\mathbb{E}_x[\Vert \sigma_{l}-\mu\Vert^2],\mu=\mathbb{E}_x[\sigma_{l}]$. $Var(\Sigma_{l})$ measures the variance of gating states of the $l$-th gating layer. $\kappa$ is a positive constant.
The above equation can also be used to quantify $H(\Sigma)$, when we simply replace $\Sigma_{l}$ with $\Sigma$.
Figure~\ref{fig:different_kappa} shows the complexity $I(X;\Sigma)$ and $I(\Sigma;Y)$, which were calculated on MLP-$\alpha$ networks learned with different $\kappa$ values on the MNIST dataset.
The $\kappa$ value affected the scale of the complexity value, but it did not affect the trend of the complexity change during the training process.
Thus, given a fixed $\kappa$ value, the complexity of different DNNs could be fairly compared.

If the DNN does not introduce additional complexity besides the input $X$, we have $H(\Sigma_{l}|X)=0$, $I(X;\Sigma_{l})=H(\Sigma_{l})-H(\Sigma_{l}|X)=H(\Sigma_{l})$. If the DNN introduces additional complexity (\emph{e.g.} using the sampling operation in VAE, or the dropout operation), then $I(X;\Sigma_{l})$ can be quantified as follows.
\begin{equation}
I(X;\Sigma_{l})\le -\frac{1}{n}\sum_{j=1}^n\log \frac{1}{n}\sum_{k=1}^n
\exp\left(-\frac{1}{2}\frac{\Vert\hat{\sigma}_{l,j}-\hat{\sigma}_{l,k}\Vert_2^2}{\sigma_0^2}\right)
\label{eq:quantify_IXS}
\end{equation}
where $\hat{\sigma}_{l,j}$ and $\hat{\sigma}_{l,k}$ represent the vectorized gating states when sampling operations are removed (in this way, we can use the method of measuring $H(\Sigma_{l})$ to quantify $I(X;\Sigma_{l})$).

Similarly, the complexity $I(X;\Sigma_{l};Y)$ can be estimated by its upper bound:
\begin{equation}
\begin{aligned}
I(X;\Sigma_{l};Y)&=I(\Sigma_l;Y)-I(\Sigma_l;Y|X)\\
&= H(\Sigma_l)-H(\Sigma_l|Y)-I(\Sigma_l;Y|X)\\
&\le
-\frac{1}{n}\sum_{j=1}^n\log \frac{1}{n}\sum_{k=1}^n
\exp\left(-\frac{1}{2}\frac{\Vert\sigma_{l,j}-\sigma_{l,k}\Vert_2^2}{\sigma_0^2}\right)\\
&\quad-\sum_{m=1}^M \! p_m\bigg[\!\!-\!\!\frac{1}{n_m}\!\!\sum_{\substack{j,\\ Y_j=m}}\!\!\!\log \frac{1}{n_m}\!\!\sum_{\substack{k,\\ Y_k=m}}\!\!\!\exp\!\left(\!\!-\frac{1}{2}\frac{\Vert\sigma_{l,j}-\sigma_{l,k} \Vert_2^2 }{\sigma_0^2}\right)\!\!\bigg]\\
&\quad-I(\Sigma_{l};Y|X)
\end{aligned}
\label{eq:quantify_IXSY}
\end{equation}
For the task of multi-category classification, $M$ denotes the number of categories. $n_m$ is the number of training samples belonging to the $m$-th category, and $p_m=n_m/M$. If the DNN does not introduce additional complexity besides the input, we have $I(\Sigma_{l};Y|X)=0$.

The entanglement of transformations is formulated as
\begin{equation}
	\begin{aligned}
    TC(\Sigma_{l})=&KL(p(\sigma_{l})||\prod_d p(\sigma_{l}^d))=\mathbb{E}_{\sigma_{l}}\left[\log \frac{p(\sigma_{l})}{\prod_d p(\sigma_{l}^d)}\right]
    \end{aligned}
\end{equation}
where $p(\sigma_{l}^d)$ denotes the marginal distribution of the $d$-th element in $\sigma_{l}$. To enable fair comparisons between $I(\hat{T},X)$ computed by the KDE method in Eq.~(\ref{eq:quantify_IXT}) and $TC(\Sigma_{l})$, we also apply the KDE method to approximate $TC(\Sigma_{l})$. To this end, we synthesize a new distribution $p(\hat{\sigma}_{l})$ to represent the distribution of $\prod_d p(\sigma_{l}^d)$.
In $\hat{\sigma}_{l}$,  $\hat{\sigma}_{l}^d$ in each dimension follows the Bernoulli distribution with the same activation rate $a_{l}^d$ with the original $\sigma_{l}^d$. Gating states $\hat{\sigma}_{l}^d$ in different dimensions are independent with each other. In this way, $\prod_d p(\sigma_{l}^d)$ can be approximated by $p(\hat{\sigma}_{l})$.

Inspired by~\cite{kolchinsky2017estimating,kolchinsky2019nonlinear}, $TC(\Sigma_{l})$ is quantified as the following upper bound.
\begin{equation}
\begin{aligned}
TC(\Sigma_{l})&=\mathbb{E}_{\sigma_{l}}\left[\log \frac{p(\sigma_{l})}{p(\hat{\sigma}_{l})}\right]\\
&\le \frac{1}{P}\sum_i\log \frac{\sum_j\exp \left(-\frac{1}{2}\frac{\Vert\sigma_{l,i}-\sigma_{l,j} \Vert^2_2}{\sigma_0^2}\right)}{\sum_j\exp \left(-\frac{1}{2}\frac{\Vert\sigma_{l,i}-\hat{\sigma}_{l,j} \Vert^2_2}{\sigma_0^2}\right)}
\end{aligned}
\label{eq:app_tc}
\end{equation}
where $P$ denotes the number of samples. $\hat{\sigma}_{l,i}$ denotes the synthesized gating states, which have the same activation rates with the gating states of the sample $i$.

\section{Learning a Minimum-Complexity DNN}
\label{sec:app_learning}
This section introduces more details about the learning of a minimum-complexity DNN in Section \ref{sec:learning} of the paper.
In Section \ref{sec:learning}, the complexity loss is defined as 
\begin{equation}
	\mathcal{L}_{\textrm{complexity}} =\sum_{l=1}^{L}H(\Sigma_{l})
	=\sum_{l=1}^{L}\{-\mathbb{E}_{\sigma_{l}}[\log p(\sigma_{l})]\}
\end{equation}
The exact value of $p(\sigma_{l})$ is difficult to calculate. Thus, inspired by~\cite{gao2018learning},  we design an energy-based model (EBM) $p_{\theta_f}(\sigma_{l})$ to approximate it, as follows.
\begin{equation}
	\begin{aligned}
	p_{\theta_f}(\sigma_{l}) &= \frac{1}{Z(\theta_f)}\exp[f(\sigma_{l};\theta_f)]\cdot q(\sigma_{l})\\
	Z(\theta_f)&=\mathbb{E}_q \left[\exp [f(\sigma_{l};\theta_f)]\right]=\int_{\sigma_{l}}q(\sigma_{l})\exp [f(\sigma_{l};\theta_f)]\mathrm{d}\sigma_{l}
	\end{aligned}
\end{equation}
where $q(\sigma_{l})$ denotes the prior distribution, which is formulated as follows.
\begin{equation}
	\begin{aligned}
	q(\sigma_{l})&=\prod_{d}q(\sigma_{l}^{d}),\quad
	q(\sigma_{l}^{d}) &=
	\begin{cases}
	\hat{p}& \sigma_{l}^{d}=1\\
	1-\hat{p}& \sigma_{l}^{d}=0
	\end{cases}
	\end{aligned}
	\label{eq:prior}
\end{equation}

If we write the EBM as $p_{\theta_f}(\sigma_{l})=\frac{1}{Z(\theta_f)}\exp[-\mathcal{E}(\sigma_{l})]$, then the energy function is as follows.
\begin{equation}
	\mathcal{E}_{\theta_f}(\sigma_{l})=-\log q(\sigma_{l})-f(\sigma_{l};\theta_f)
\end{equation}

The EBM can be learned via the maximum likelihood estimation (MLE) with the following loss.
\begin{equation}
	\hat{\theta}_f =  \mathop{\arg\max}_{\theta_f}	L(\theta_f) = \mathop{\arg\max}_{\theta_f} \frac{1}{n}\sum_{i=1}^{n}\log p_{\theta_f}(\sigma_{l,i})
	\label{equ:EBM_{l}earning}
\end{equation}
where $n$ denotes the number of samples. $\sigma_{l,i}$ is a vector, which represents gating states in the $l$-th gating layer for the $i$-th sample.

The loss and gradient of $\theta_f$ can be calculated as follows.
\begin{equation}
	L(\theta_f)=-\frac{1}{n}\sum_{i=1}^n\log p_{\theta_f}(\sigma_{l,i})
	=-\frac{1}{n}\sum_{i=1}^n[f(\sigma_{l,i};\theta_f)+\log q(\sigma_{l,i})]+\log Z(\theta_f)
\end{equation}

\begin{equation}
	\frac{\partial L(\theta_f)}{\partial\theta_f}=\mathbb{E}_{\theta_f}\left[\frac{\partial}{\partial \theta_f}f(\sigma_{l};\theta_f)\right]-\frac{1}{n}\sum_{i=1}^n\frac{\partial}{\partial\theta_f}f(\sigma_{l,i};\theta_f)
\end{equation}
where $\frac{\partial}{\partial\theta_f}\log Z(\theta_f) = \mathbb{E}_{\theta_f}[\frac{\partial}{\partial\theta_f}f(\sigma_{l};\theta_f)]$.

The first term $\mathbb{E}_{\theta_f}\left[\frac{\partial}{\partial \theta_f}f(\sigma_{l};\theta_f)\right]$ in the above equation is analytically intractable and has to be approximated by MCMC, such as the Langevin dynamics. 
\begin{equation}
	\begin{aligned}
	\sigma_{l}^{\text{new}}&=\sigma_{l}-\frac{\Delta \tau}{2}\frac{\partial}{\partial\sigma_{l}}\mathcal{E}_{\theta_f}(\sigma_{l})+\sqrt{\Delta \tau}\epsilon\\
	&= \sigma_{l}+\frac{\Delta\tau}{2}\left[{\frac{\partial f(\sigma_{l};\theta_f)}{\partial \sigma_{l}}+\sum_{d=1}^D\frac{1}{q(\sigma_{l}^{d})}\frac{\partial q(\sigma_{l}^{d})}{\partial \sigma_{l}^{d}}}\right]+\sqrt{\Delta \tau}\epsilon
	\end{aligned}
\end{equation}
where $\epsilon\backsim N(\mathbf{0},\mathbf{I})$ is a Gaussian white noise. $\Delta \tau$ denotes the size of the Langevin step.

Then, the Monte Carlo approximation to $\frac{\partial L(\theta_f)}{\partial \theta_f}$ is given as follows.
\begin{equation}
	\begin{aligned}
	\frac{\partial L(\theta_f)}{\partial\theta_f}&\approx \frac{1}{n}\sum_{i=1}^{n}\frac{\partial}{\partial\theta_f}f(\widetilde{\sigma}_{l,i};\theta_f)-\frac{1}{n}\sum_{i=1}^n\frac{\partial}{\partial\theta_f}f(\sigma_{l,i};\theta_f)\\
	&=\frac{\partial}{\partial\theta_f}\left[\frac{1}{n}\sum_{i=1}^n\mathcal{E}_{\theta_f}(\sigma_{l,i})-\frac{1}{n}\sum_{i=1}^n\mathcal{E}_{\theta_f}(\widetilde{\sigma}_{l,i})\right]
	\end{aligned}
	\label{eq:mcapproxL}
\end{equation}
where $\widetilde{\sigma}_{l,i}$ is the sample synthesized via Langevin dynamics.

Thus, the loss for the learning of the DNN can be rewritten as follows.
\begin{equation}
\begin{aligned}
\mathcal{L}_{\textrm{complexity}}&=-\sum_{l=1}^{L}\mathbb{E}_{\sigma_{l}}[\log p_{\hat{\theta}_f}(\sigma_{l})]\\
&=\frac{1}{n}\sum_{l=1}^{L}\sum_{i=1}^n[\mathcal{E}_{\hat{\theta}_f}(\sigma_{l,i})+\log Z(\hat{\theta}_f)]\\
&=-\frac{1}{n}\sum_{l=1}^{L}\sum_{i=1}^n[f(\sigma_{l,i};\hat{\theta}_f)+\log q(\sigma_{l,i})-\log Z(\hat{\theta}_f)]
\end{aligned}
\label{equ:loss_tractable}
\end{equation}

Let $\theta_{\textrm{DNN}}$ denote parameters in the DNN. The gradient of $\theta_{\textrm{DNN}}$ can be calculated as follows.
\begin{equation}
	\begin{aligned}
    \frac{\partial Loss}{\partial\theta_{\textrm{DNN}}} =&-\frac{1}{n}\sum_{l=1}^{L}\sum_{i=1}^n\left\lbrace\frac{\partial f(\sigma_{l};\hat{\theta}_f)}{\partial \sigma_{l,i}} +\sum_d^D\frac{1}{q(\sigma_{l,i}^{d})}\frac{\partial q(\sigma_{l,i}^{d})}{\partial \sigma_{l,i}^{d}}\right\rbrace \frac{\partial \sigma_{l,i}}{\partial \theta_{\textrm{DNN}}}
    \end{aligned}
\end{equation}
We consider $Z(\theta_f)$ as a constant in the computation of $\frac{\partial Loss}{\partial \hat{\theta}_{\textrm{DNN}}}$.

To enable the computation of $\frac{\partial\sigma_{l,i}}{\partial\theta_{DNN}}$ and $\frac{\partial q(\sigma_{l,i}^{d})}{\partial\sigma_{l,i}^{d}}$, we can approximate the ReLU operation using the following Swish function~\cite{ramachandran2017searching}.
\begin{equation}
\begin{aligned}
\sigma_{l} &\approx \text{sigmoid}(\beta x)\\
\text{ReLU}(x) & =x \odot \sigma_{l} \approx x \odot \text{sigmoid}(\beta x)
\end{aligned}
\end{equation}
where $\odot$ denotes the element-wise multiplication.

According to Eq.~(\ref{eq:prior}), the prior distribution $q(\sigma_{l})$ is approximated as follows.
\begin{equation}
    q(\sigma_{l}^{d}) \approx 1-\hat{p} + \sigma_{l}^{d}(2\hat{p}-1), \quad \frac{\partial q(\sigma_{l}^{d})}{\partial \sigma_{l}^{d}} \approx 2\hat{p}-1
\end{equation}
In implementation, the EBM is a bottom-up ConvNet with six convolutional layers, which takes $\sigma_{l}$ as an input and outputs a scalar.
During the training phase, we firstly train the EBM using Eq.~(\ref{eq:mcapproxL}) for every batch of training data.
The EBM and the original DNN are trained separately. \emph{I.e.} when training the EBM, parameters in the original DNN are fixed, and vice versa.

\section{More Experimental Details, Results, and Discussions}

\subsection{More experimental details}
\label{sec:app_samples}

Recall that $\Sigma_l=\{\sigma_l\}$ denotes the set of gating states $\sigma_l$ among all samples $X$. In this paper, we randomly sample 2000 images from the training set of the each dataset for the calculation of the transformation complexity. Thus, $X$ denotes the set of 2000 randomly sampled images that are used for analysis.

\subsection{The value of $\kappa$ used in the KDE approach}
\label{sec:app_kappa}
In this section, we discuss about the value of the hyper-parameter $\kappa$ used in the KDE approach.
Note that the features of convolutional layers usually contain far more dimensions than features of fully-connected layers.
Therefore, we set $\kappa=0.04$ for gating layers following each convolutional layer, and $\kappa=0.01$ for gating layers following each FC layer.

We also tested the effects of different $\kappa$ values to the quantification of the complexity.
Figure~\ref{fig:different_kappa} shows the complexity $I(X;\Sigma)$ and $I(\Sigma;Y)$, which were calculated on MLP-$\alpha$ networks learned with different $\kappa$ values on the MNIST dataset.
The $\kappa$ value affected the scale of the complexity value, but it did not affect the trend of the complexity change during the training process.
Thus, given a fixed $\kappa$ value, the complexity of different DNNs could be fairly compared.

\subsection{The classification accuracy of DNNs in comparative studies}
\label{sec:app_acc}
This section contains more details of DNNs in comparative studies of the paper. We trained five types of DNNs on the MNIST dataset and the CIFAR-10 dataset, and trained three types of DNNs on the CelebA dataset and the Pascal VOC 2012 dataset. Table~\ref{tab: acc_loss} reports the testing accuracy of the trained DNNs.
\vspace{-10pt}
\begin{table}[h]
\centering
\caption{The classification accuracy of DNNs on different datasets.\vspace{3pt}}
\begin{minipage}{.58\linewidth}
\centering
{\small (a) On the MNIST and CIFAR-10 datasets.}
\vfill
\resizebox{\textwidth}{!}{
\begin{tabular}{c| c c c c c}
	\hline
	{} & MLP & LeNet-5  & revised VGG-11 & ResNet-20 & ResNet-32\\
	\hline
	MNIST & 96.52\% & 97.41\% & 99.00\% & 98.70\% & 98.24\% \\
	CIFAR-10 & 52.52\% & 61.5\%  & 84.53\% & 81.75\% & 79.76\% \\
	\hline				
\end{tabular}
}
\centering
\vspace{3pt}
\end{minipage}
\hfill
\begin{minipage}{.40\linewidth}
\centering
{\small (b) On the CelebA and Pascal VOC 2012 datasets.}
\vfill
\resizebox{\textwidth}{!}{
\begin{tabular}{c| c c c }
	\hline
	{} &  ResNet-18 & ResNet-34 & VGG-16 \\ \hline
	CelebA & 80.25\% & 80.91\% & 89.70\% \\
	Pascal VOC 2012 &  67.99\% & 64.27\% & 62.50\% \\
	\hline			
\end{tabular}
}
\vspace{3pt}
\end{minipage}
\label{tab: acc_loss}
\end{table}

\subsection{The change of the transformation complexity in VGG-16}
\label{sec:vgg_complexity}
\begin{figure}[h]
	\centering
	\includegraphics[width=0.6\linewidth]{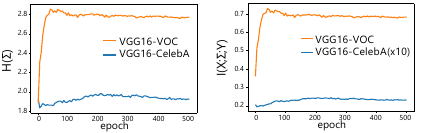}
	\vspace{-10pt}
	\caption{The change of transformation complexity in VGGs.}
	\vspace{-5pt}
	\label{fig:change_vgg}
\end{figure}

This section shows more results of the change of transformation complexity in Section \ref{sec:exp}.
Figure~\ref{fig:change_vgg} shows the change of transformation complexity in  VGG-16 trained on the Pascal VOC
dataset and the CelebA dataset. In these cases, the complexity increased monotonously during the early stage of the training process, and saturated later.

\subsection{More experimental result of learning DNNs with minimum complexity}

\begin{figure}[htbp]
\centering
\begin{minipage}{.55\linewidth}
\centering
\includegraphics[width=\linewidth]{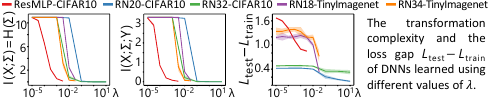}
\includegraphics[width=\linewidth]{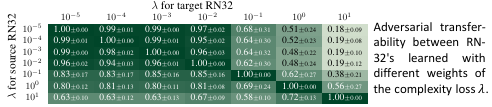}
\vspace{-20pt}
\caption{(Top) The complexity of transformation complexity and the gap between the training loss and the testing loss of the learned minimum complexity DNNs, with the standard deviation in shaded colors. (Bottom) Adversarial transferability between DNNs learned with different weights of the complexity loss {\small$\lambda$}, with standard deviations.}
\label{fig:rebuttal-exp}
\vspace{-5pt}
\end{minipage}
\hfill
\begin{minipage}{.42\linewidth}
\centering
\includegraphics[width=.95\linewidth]{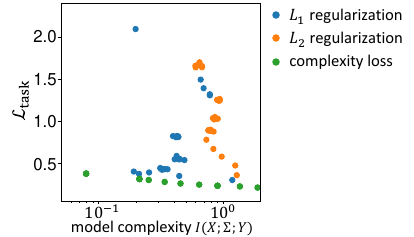}
\vspace{-10pt}
\caption{Comparisons of the classification loss between different baselines. The complexity loss was superior to them in terms of maintaining the classification accuracy and decreasing the model complexity.}
\vspace{-5pt}
\label{fig:compare-l1-l2}
\end{minipage}
\end{figure}

In this section, we provide more experimental result to verify the stability of learning DNNs with minimum complexity in Section \ref{sec:learning}.
Specifically, we repeated experiments in Figure \ref{fig:minimum} and Figure \ref{fig:robustness-transferability}(b) for six times with different random initializations.
Results in Figure \ref{fig:rebuttal-exp} with the standard deviation still verify our conclusions.

\subsection{Comparisons with traditional $L_1$ and $L_2$ regularization methods}
\label{subsec:app_compare_l1_l2}

In this section, we compare the proposed complexity loss with traditional $L_1$ and $L_2$ regularization methods.
Specifically, we trained 20 residual MLPs with $\mathcal{L}=\mathcal{L}_{\text{task}}+\lambda_{L_1}\sum_l \Vert W_l\Vert_1$ and 20 residual MLPs with $\mathcal{L}=\mathcal{L}_{\text{task}}+\lambda_{L_2}\sum_l \Vert W_l\Vert_2^2$.
For the $L_1$ regularization, we set different values of $\lambda_{L_1}$ ranging from $10^{-5}$ to $10^{-1}$ in different experiments.
For the $L_2$ regularization, we set different values of $\lambda_{L_2}$ ranging from $10^{-5}$ to $10^1$ in different experiments.
Figure \ref{fig:compare-l1-l2} compares the classification loss $\mathcal{L}_{\text{task}}$ between different baselines.
We found that the complexity loss was superior to traditional $L_1$ and $L_2$ regularization methods, in terms of maintaining the classification accuracy and decreasing the model complexity.

\subsection{Detailed analysis of the adversarial robustness, adversarial transferability and knowledge consistency}
\label{subsec:app_detail_robust_trans_consis}

In this section, we provide more detailed analysis of the adversarial robustness, adversarial transferability and knowledge consistency, in Section \ref{sec:learning} of the paper.

In Section \ref{sec:learning} of the paper, we have found that DNNs with low transformation complexity usually exhibited high adversarial robustness; vice versa.
To this end, we defined the attacking utility in untargeted attacks as $U_{\text{untarget}}(x)=\max_{y'\neq y}h_{y'}(x+\epsilon)-h_{y}(x+\epsilon)$, where $y$ is the ground-truth label of the sample $x$, and $h_y(x)$ is the output logit of the DNN in the $y$-th category given an input $x$. Then, we measured the $L_2$ norm of the minimum adversarial perturbation $\epsilon$ for each image, which has similar attacking utility of 40.
We conducted the PGD attack~\cite{madry2017towards}, with the step size of each single-step attack as $0.5/255$. Other experimental settings
remained the same as in~\cite{wang2020unified}.

We have also found in Section \ref{sec:learning} that adversarial perturbations for complex DNNs could not be well transferred to simple DNNs. However, adversarial perturbations for simple DNNs could be transferred to complex DNNs. We conducted the PGD attack for each image using DNNs with different transformation complexities, and following the experimental settings in~\cite{wang2020unified} to measure the adversarial transferability between DNNs.

Following settings in~\cite{liang2019knowledge}, we explored the knowledge consistency between DNNs with different transformation complexities.
Specifically, we used the intermediate-layer feature $x_A$ of a DNN (\textit{Net-A}) trained with a specific value of $\lambda$, to reconstruct the intermediate-layer feature $x_B$ of another DNN (\textit{Net-B}) also trained with $\lambda$. \textit{Net-A} and \textit{Net-B} had the same architecture and were trained on the same dataset, but with different initialization parameters. $x_A$ and $x_B$ denote intermediate-layer features of \textit{Net-A} and \textit{Net-B} in the same layer, respectively.

We followed the experimental settings in~\cite{liang2019knowledge} to diagnose the feature representation in the residual MLP trained on the CIFAR-10 dataset and ResNet-34 trained on the Tiny-ImageNet dataset. We diagnosed the output feature of the last layer ($3072$ dimensional) in the residual MLP and the output feature of the last convolutional layer ($7\times7\times512$ dimensional) in ResNet-34. We disentangled 0-order, 1-order, and 2-order consistent features $x^{(0)}$, $x^{(1)}$, and $x^{(2)}$ from $x_A$. For the fair comparison between DNNs learned with different $\lambda$ values, we computed the strength of the $k$-order consistent feature as $Var(x^{(k)})/Var(x_A)$. $Var(x_A)\triangleq\mathbb{E}_{I,i}[(x_{A,I,i}-\mathbb{E}_{I',i'}[x_{A,I',i'}])^2]$, where $x_{A,I,i}$ denotes the $i$-th element of $x_A$ given the image $I$.

\begin{figure}[h]
    \centering
    \includegraphics[width=.25\linewidth]{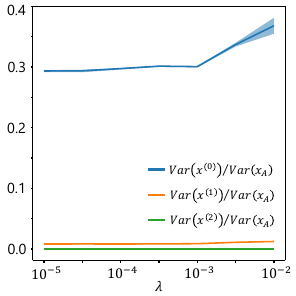}
    \includegraphics[width=.25\linewidth]{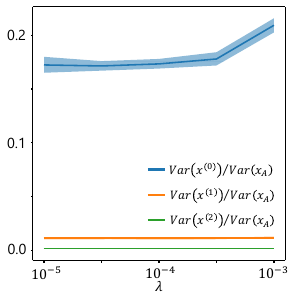}
    \vspace{-1em}
    \caption{The strength of consistent features with different orders in (left) residual MLP trained on the CIFAR-10 dataset, and (right) ResNet-34 trained on the Tiny ImageNet dataset.}
    \label{fig:knowledge-consistency}
\end{figure}

Figure~\ref{fig:knowledge-consistency} shows the strength of consistent features with different orders.
We found that pairs of simple DNNs usually encoded similar knowledge representations (exhibiting high knowledge consistency), while complex DNNs are more likely to encode diverse knowledge.
This demonstrated the reliability of features learned by simple DNNs.

\subsection{Computational cost of the proposed method}
\label{subsec:app_computational_cost}
This section reports the computational cost of training DNNs with the complexity loss. We compare the time cost of training for an epoch with and without the complexity loss.
The time cost was measured using PyTorch 1.6 \cite{pytorch} on Ubuntu 18.04, with the Intel(R) Core(TM) i9-10900X CPU @ 3.70GHz and one NVIDIA(R) TITAN RTX(TM) GPU.
\vspace{-10pt}
\begin{table}[h]
	\centering
	\caption{Time cost of training DNNs for an epoch with and without the complexity loss.\vspace{3pt}}
	\resizebox{.7\textwidth}{!}{%
	\begin{tabular}{c| c c c c}
		\hline
		model &  dataset & batch size & time w/o $L_{\textrm{complexity}}$ & time w/ $L_{\textrm{complexity}}$ \\
		\hline
		residual MLP & CIFAR-10 & 128 & 21 s/epoch &  85 s/epoch\\
		ResNet-20 & CIFAR-10 & 128 &  64 s/epoch &  218 s/epoch\\
		ResNet-32 & CIFAR-10 & 128 &  101 s/epoch &  296 s/epoch\\
		ResNet 18 & Tiny-ImageNet & 128 &  41 s/epoch &  83 s/epoch\\
		ResNet-34 & Tiny-ImageNet & 128 &  68 s/epoch &  105 s/epoch\\
		\hline			
	\end{tabular}
	}
\end{table}

\end{document}